%% file: root.tex
\documentclass[letterpaper, 10 pt, conference]{ieeeconf}  

\IEEEoverridecommandlockouts                              

\usepackage{graphics} 
\usepackage{amsmath} 
\usepackage{tabularx}
\usepackage{booktabs}
\usepackage{algorithm}
\usepackage{algorithmic}
\usepackage{subcaption}
\usepackage[dvipsnames]{xcolor}
\usepackage{graphicx}
\graphicspath{ {./imgs/} }
\usepackage{hyperref}

\title{Non Holonomic Collision Avoidance under Non-Parametric Uncertainty: A Hilbert Space Approach}
\author{Unni Krishnan R Nair*$^1$, Anish Gupta*$^1$,  D. A. Sasi Kiran$^1$, Ajay Shrihari$^1$, Vanshil Shah$^1$,\\ 
Arun Kumar Singh$^2$, K. Madhava Krishna$^1$ 
\thanks{1. Robotics Research Center, IIIT Hyderabad, India. 2.  Institute of Technology, University of Tartu. * Equal Contribution}
}

\newcommand{\norm}[1]{\left\lVert#1\right\rVert}
\DeclareMathOperator*{\argmin}{arg\,min}

\begin{document}

\maketitle
\thispagestyle{empty}
\pagestyle{empty}

\input{chapters/abstract}
\input{chapters/introduction}
\input{chapters/related-work}
\input{chapters/preliminaries}
\input{chapters/methodology}

\input{chapters/Sim1}

\input{chapters/conclusion}
\bibliographystyle{IEEEtran}
\bibliography{references} 

\end{document}

%% file: chapters/abstract.tex
\begin{abstract}

We consider the problem of an agent/robot with non-holonomic kinematics avoiding dynamic and static obstacles. Additionally there may be bounds/constraints on the configurational space of the robot in the form of lane/corridor boundaries. State and velocity noise of the robot, the lanes, the obstacles, and the robot's control noise are modelled as non-parametric distributions as Gaussian assumptions of noise models are violated in real-world scenarios. Under these assumptions, we formulate a robust MPC that samples robotic controls effectively in a manner that aligns the robot to the goal state while avoiding obstacles and staying within the lane bounds under the duress of such non-parametric noise. In particular, the MPC incorporates a distribution matching cost that effectively aligns the distribution of the current collision cone to a certain desired distribution whose samples are collision-free. This cost is posed as a distance function in the Hilbert Space, whose minimization typically results in the collision cone samples becoming collision-free. We show tangible performance gains compared to methods that model the collision cone distribution by linearizing the Gaussian approximations of the original non-parametric state and obstacle distributions. We also show superior performance to methods that pose a chance constraint formulation of the Gaussian approximations of non-parametric noise without subjecting such approximations to further linearizations. The performance gain is shown both in terms of trajectory length and control costs that vindicates the efficacy of the proposed method. Finally we show the proposed method being used to navigate with a non holonomic differential drive robot in real-time in a realistic setting in Gazebo with dynamic and static obstacles. To the best of our knowledge, this is the first presentation of non-holonomic collision avoidance of stationary obstacles, moving obstacles and lane constraints in the presence of non-parametric state, velocity, actuator and lane boundary noise models.
\end{abstract}

%% file: chapters/introduction.tex
\section{Introduction}

Collision avoidance of dynamic obstacles by kinematically constrained planar robots has been a problem well studied in literature \cite{Mora-ICRA14, Manocha-GVO}. However, collision avoidance in the presence of uncertainty of such systems has been sparsely reported. All the more so, when state and control noise (that includes velocity and actuator noise) are characterized by non-parametric models, which is typically encountered in real-world and practical scenarios. Recently \cite{pvo-ral2020, gopalakrishnan2018tcst} details collision avoidance, as well as trajectory tracking under the duress of non-parametric noise but wherein the kinematic evolution can be expressed in the form of linear models. In this paper, we propose a formulation to kinematically constrained systems wherein the state evolves according to the popular unicycle kinematics \cite{unicycle}.

\begin{figure}
    \includegraphics[width=.49\textwidth]{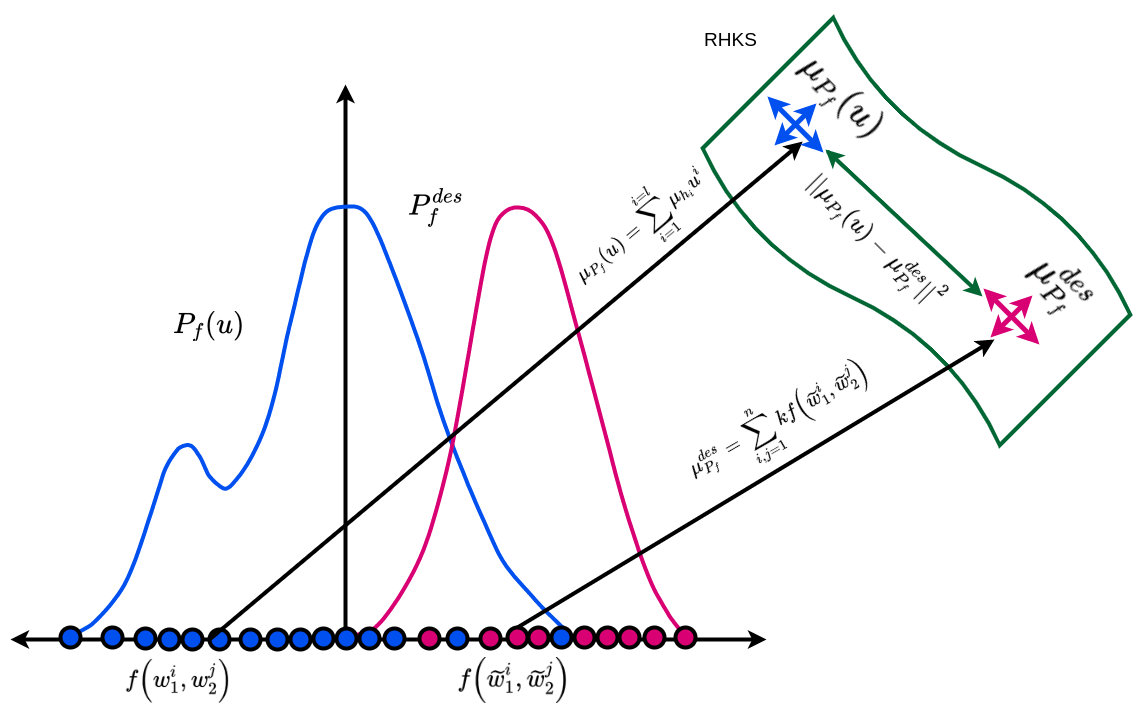}\hfill
    \caption{\small Probability density functions in the physical space can be represented as functions (or points) in RKHS and therefore the similarity between them can be expressed as distance between points. Here the distributions $P_f({\textbf{u}})$ and $P_f^{des}$ are parameterized by $w_1^{i}$, $w_2^{i}$ and $\widehat{w_1}^{i}$, $\widehat{w_2}^{i}$.\normalsize}
    \label{RKHS-Embed}
\end{figure}

This paper proposes a novel MPC formulation that outputs actuator controls in the form of changes in linear and angular velocities when subject to non-parametric state, velocity, control and lane boundary noise of both the ego agent and other dynamic/static participants. The MPC formulation takes the form of matching the moments of a certain desired collision cone distribution (whose samples are all collision-free) and the current distribution of the ego agent's collision cone vis-a-vis other dynamic participants. We also define a desired distribution for velocity samples that avoid the lane boundaries , the MPC formulation also tries to match the moments of this distribution to the current distribution due to ego agent's current state. By collision cone, we indicate the analytical counterpart of the popular velocity obstacle \cite{vo} that has been successfully employed in many works \cite{Mora-ICRA14, manocha-rvo, prvo}. The collision cone (\ref{eq:vo}) distribution is non-parametric as well since it is composed of agent and obstacle state and velocities whose distributions are non-parametric. The matching of higher-order moments of two non-parametric distributions can be accomplished by embedding these distributions into the Reproducing Kernel Hilbert Space (RKHS) \cite{scholkopf, scholkopf2}, as shown in Figure \ref{RKHS-Embed} through the Kernel trick. Those actuator controls that minimize this distance in the RKHS space are selected by the proposed MPC formulation from a set of feasible controls.

In particular the paper contributes in the following manner 
\begin{enumerate}
    \item It proposes a novel method utilizing RKHS based collision avoidance \cite{pvo-ral2020} for kinematically constrained robots/agents whose state evolution can be modelled by unicycle kinematics under the duress of non-parametric noise. Typically the dynamic obstacle avoidance problem is solved for a holonomic robot and then a non-holonomic controller is used to best track this trajectory. We circumvent this by sampling achievable velocities in the $(v,w)$ space of the non holonomic robot, and then choosing an optimum sample.
    \item We also provide a framework for collision avoidance in a realistic setting by bringing in lane/corridor constraints with non parametric noise which provide boundaries to the robots configuration space as typical work-spaces are always bounded. We formulate lane keeping also into a distribution matching problem similar to our obstacle avoidance problem stated above bringing in a unified approach to deal with non parametric uncertainty.
    \item Moreover, the RKHS formulation provides for an easily tunable parameter through which a trade-off between cost and robustness is possible. We showcase this through ablations on this parameter in the results section.
\end{enumerate}

Favourable comparisons in terms of trajectory length, control costs as well as qualitative evaluations vis-a-vis many dynamic participants vindicate the efficacy of the proposed framework in Sections \ref{sec:Ablation} and \ref{sec:multiobs} respectively. It is to be noted that performance gain over other distribution matching frameworks such as KLD was established in our previous works \cite{pvo-ral2020}. The proposed method also compares favourably against two variants which approximates the non-parametric noise through Gaussian models, one which linearizes the Gaussian approximations of state and velocities of all participants to obtain linearized collision cone distribution, the other which does not subject the Gaussian models to such linearizations. Both these variants are posed as Chance Constraint Optimization along the lines reported in \cite{iros15_bharath, prvo}. We further also showcase real-time performance in the control of a non holonomic robot in a realistic gazebo setting under the duress of non parametric noise.

\label{sec:introduction}

%% file: chapters/related-work.tex
\section{Related Work}
While literature abounds in collision avoidance formulations of dynamic obstacles from both single-agent and multi-agent standpoints \cite{vo, manocha-rvo} the literature is sparse when it comes to collision avoidance under uncertainty. The trade-off comes in the form of just avoiding the mean evolution of dynamic obstacles with heightened collision chances vis-a-vis avoiding the entire uncertainty distribution that gives rise to highly conservative maneuvers. Earlier methods tended to err on the side of caution by growing obstacles by the size of the uncertainty, often referred to as bounding volume methods \cite{bounding_volume1, bounding_volume2}. However, the limitations of such approaches were exemplified in \cite{prvo} where it was shown chance-constrained formulations offer tunable trade-offs. 

Posing the problem as a robust MPC with probabilistic chance constraints are by nature intractable. The primary difficulty lies in computing the analytical form for the chance constraints. Notable exceptions exist in the case in which the random variables under consideration have Gaussian distribution and the chance constraints are defined over affine inequalities \cite{boyd_chance, chance_blackmore}. Formulations along the lines of \cite{Bratz, Kothari, prvo} have developed reliable surrogates that give tight approximations to these chance constraints defined over non-linear inequalities(could be non-convex also). However, all these algorithms require a fundamental assumption on the nature of uncertainty of the random variables (state and actuation of the robot) involved. In general closed-form, surrogates can only be derived if the random variables belong to a Gaussian distribution. Extensions to the non-Gaussian and non-parametric case are very complex and a very active field of research.

The challenge of robust MPC in the presence of non-parametric uncertainty for collision avoidance systems was first approached in \cite{pvo-ral2020} wherein an RKHS embedding of moment matching cost was folded into the original MPC cost function. These approaches were inspired by the formulations of \cite{scholkopf, scholkopf2} and showcased their efficacy on systems whose state evolution was linear with respect to the current state. 

%% file: chapters/preliminaries.tex
\section{Preliminaries}
\label{sec:preliminaries}

In this paper, we represent the vectors in bold letters, $\textbf{x}$, matrices in capital, $M$, and sets in mathcal, $\mathcal{A}$. $\norm{\textbf{x}}$ denotes the Euclidean norm of $\textbf{x}$. For a random variable $\text{x}$, $\hat{\text{x}}$ denotes the mean. $\Pr(\cdot)$ denotes the probability of an event while $P(\cdot)$ denotes the probability density function. The subscript $\cdot_t$ indicates the value at time $t$ while the superscript $\cdot^j$ indicates the $j^{th}$ robot or obstacle. Some of the commonly used symbols and notations are summarized in the table \ref{tab:notations}. We also define some notations in the first place of their use.

\begin{table}
\caption{}
\label{tab:notations}
\begin{tabular}{@{}ll@{}}
\toprule
\multicolumn{1}{c}{\textbf{Symbol}} & \multicolumn{1}{c}{\textbf{Description}} \\ \midrule
$(^r\textbf{x}_t, ^r\dot{\textbf{x}}_t)$ & (Position, Velocity) of the robot at time $t$ \\ \midrule
$(\theta_t, \omega_t)$ & (Heading, Angular velocity) of the robot at time $t$ \\ \midrule
$(^o\textbf{x}^j_t, ^o\dot{\textbf{x}}^j_t)$ & (Position, Velocity) of the $j^{th}$ obstacle at time $t$ \\ \midrule
$\textbf{u}_{t}$ & Control input to the robot at time $t$ \\ \midrule
$f^j_t(\cdot) \leq 0$ & Deterministic collision avoidance constraint \\ \midrule
$P(f^j_t(\cdot))$ & Distribution of $f^j_t(\cdot)$ under uncertainty \\ \midrule
$\eta$ & Probability of satisfaction of the chance constraint \\ \bottomrule
\end{tabular}
\end{table}

\subsection{Robot State Estimation}
\label{sec:robot-state-estimation}
The state estimates of the robot at time $t$ is given by $(^r\textbf{x}_t, ^r\dot{\textbf{x}}_t)$. We model $^r\textbf{x}_t$ and $\dot{^r\textbf{x}}_t$ as random variables following a non parametric probability distribution. We define the state evolution over time in (\ref{eq:state-evolution}).


\begin{subequations}
    \label{eq:state-evolution}
    \begin{equation}
        ^r\textbf{x}_t = A^r\textbf{x}_{t-1} + B(^r\dot{\textbf{x}}_{t})
    \end{equation}
    \begin{equation}
        ^r\dot{\textbf{x}}_{t} = \begin{bmatrix}
            v_t \cos(\theta_{t-1} + \omega_t \Delta t)\\
            v_t \sin(\theta_{t-1} + \omega_t \Delta t)
        \end{bmatrix}
    \end{equation}
    \begin{equation}
        \textbf{u}_t + \delta_t = \begin{bmatrix}
            v_t \\
            \omega_t
        \end{bmatrix}
    \end{equation}
\end{subequations}


where, $\delta_t$ is a random variable described by an unknown non-parametric probability distribution and denotes the uncertainty in executing the given control. $v_t$ is the magnitude of the velocity of the robot at time $t$ while $\omega_t$ is the angular velocity. $\theta_t$ is the heading of the robot at time $t$. Though the distribution of $\delta_t$ is unknown, we assume that we have access to the samples of $\delta_t$ at every instance of $t$. We apply a particle filter-based approach on (\ref{eq:state-evolution}) to propagate the noise in time space.

\subsection{Obstacle State Estimation}
\label{sec:obstacle-state-estimation}
We assume that we have access to the state estimates $^o\textbf{x}^j_t$ of the obstacles via a sensor like a camera or a LiDAR. The sensors are prone to noise and hence $^o\textbf{x}^j_t$ is also a random variable described by an unknown non-parametric probability distribution. We estimate $^o\dot{\textbf{x}}^j_t = D(^o\textbf{x}^j_t)$ using another particle filter to track the obstacles.

\subsection{Collision Avoidance Chance Constraints}
\label{sec:collision-avoidance-chance-constraints}
\subsubsection{Collision avoidance}
Consider the ego and obstacle modeled as circular disks with radii $^rR$ and $^oR^j$ respectively. We express the collision avoidance constraint at a given time $t$ based on the velocity obstacle method [6] in (\ref{eq:vo}).

\begin{subequations}
    \label{eq:vo}
    \begin{equation}
        f^j_t(\cdot) \leq 0 : \frac{(\textbf{r}_j^T \textbf{v}_j)^2}{\norm{\textbf{v}_j}^2} - \norm{\textbf{r}_j}^2 + (^rR + ^oR^j)^2 \leq 0, \forall j
    \end{equation}
    \begin{equation}
        \textbf{r}_j = {^r\textbf{x}_t} - {^o\textbf{x}^j_t},\ \ 
        \textbf{v}_j = {^r\dot{\textbf{x}}_t} - {^o\dot{\textbf{x}}^j_t}
    \end{equation}
\end{subequations}

\maketitle

\subsubsection{Chance Constraints}
\label{sec:pvo}
According the state estimations described in sections \ref{sec:robot-state-estimation} and \ref{sec:obstacle-state-estimation}, the positions and velocities of the robot and obstacles are random variables described by unknown probability distributions. Hence, the collision avoidance constraints described in (\ref{eq:vo}) should be satisfied in a probabilistic manner given by (\ref{eq:pvo}).

\begin{equation}
    \label{eq:pvo}
    \Pr(f^j_t(\cdot) \leq 0) > \eta, \forall j
\end{equation}

Ie. a control $\textbf{u}$ is required which moves the distribution to the left of zero. Figure \ref{distribution-shift} shows how the shape of the distribution can be manipulated by $\textbf{u}$ to obtain a desired distributional shift.

\subsection{Corridor/lane Constraints}
\subsubsection{Keeping within the corridor}
\label{sec:clo}
The corridor is defined as two linear boundaries within which the robot has to be in at all times. The corridor is defined as 2 lines 
\begin{equation}
    c_{1}ax+c_{1}bx+c_{1}c = 0
\end{equation}
\begin{equation}
    c_{2}ax+c_{2}bx+c_{2}c = 0
\end{equation}
the collision avoidance condition at time t to avoid collision with the corridor boundary is defined as:
\begin{subequations}
    \label{eq:clo}
    \begin{equation}
        {d_{1}}^j_t(\cdot) \leq 0 : (c_{1}a\textbf{x}_{t+1}[0]+c_{1}b\textbf{x}_{t+1}[1]+c_{1}c)/\sqrt{c_{1}a^{2}+c_{1}b^{2}} \leq 0
    \end{equation}
    \begin{equation}
        {d_{2}}^j_t(\cdot) \geq 0 : (c_{2}a\textbf{x}_{t+1}[0]+c_{2}b\textbf{x}_{t+1}[1]+c_{2}c)/\sqrt{c_{2}a^{2}+c_{2}b^{2}} \geq 0
    \end{equation}
\end{subequations}

\subsubsection{Chance Constraints for corridor}
\label{sec:pclo}
The collision avoidance constraints described in (\ref{eq:clo}) should be satisfied in a probabilistic manner given by (\ref{eq:pclo}).

\begin{subequations}
    \label{eq:pclo}
    \begin{equation}
    \Pr({d_{1}}^j_t(\cdot) \leq 0) > \eta_{corr1}, \forall j
    \end{equation}
    \begin{equation}
    \Pr({d_{2}}^j_t(\cdot) \geq 0) > \eta_{corr2}, \forall j
    \end{equation}
\end{subequations}

\subsection{Problem Formulation}
\label{sec:problem-formulation}
We formulate the collision avoidance problem as an optimization problem. For every obstacle, we satisfy the chance constraints while optimizing a trajectory following cost.

\begin{subequations}
    \label{eq:mpc}
    \begin{equation}
        \label{eq:mpc-cost}
        \argmin_{u_t} J(\textbf{u}_t) = \norm{^r\hat{\textbf{x}}_t - ^{d}\textbf{x}_t}^2 + \norm{u_t}^2
    \end{equation}
    \begin{equation}
        \label{eq:mpc-chance-constraints}
        \text{subject to} \ \Pr(f^j_t(\cdot) \leq 0) > \eta,\ \forall j,\  \textbf{u}_t \in \mathcal{C}
    \end{equation}
        \begin{equation}
        \label{eq:mpc-lane1-chance-constraints}
        \text{subject to} \ \Pr({d_{1}}^j_t(\cdot) \leq 0) > \eta_{corr1}, \forall j,\  \textbf{u}_t \in \mathcal{C}
    \end{equation}
        \begin{equation}
        \label{eq:mpc-lane2-chance-constraints}
        \text{subject to} \ \Pr({d_{2}}^j_t(\cdot) \geq 0) > \eta_{corr2}, \forall j,\  \textbf{u}_t \in \mathcal{C}
    \end{equation}
\end{subequations}

where, $^d\textbf{x}_t$ is some desired velocity that the robot needs to attain at time $t$ and $\mathcal{C}$ is the set of feasible control commands.

%% file: chapters/methodology.tex
\section{Methodology}
\label{sec:methodology}


We now present the method to sample the set of feasible controls, $\mathcal{C}$, for a non-holonomic robot. We also describe a method to solve the chance constraints presented in (\ref{eq:pvo}). In general, the chance constraint of the form (\ref{eq:mpc-chance-constraints}) is intractable and is replaced by surrogate constraints. For non-parametric state and control noise, we model the surrogate constraint as distribution matching problem \cite{pvo-ral2020}, \cite{gopalakrishnan2018tcst}. From this, we reformulate (\ref{eq:mpc}) as

\begin{equation}
    \begin{aligned}
    \label{eq:mpc-rkhs}
    \argmin_{u_t} J(\textbf{u}_t) + \lambda \mathcal{L}_{dist}(P(f^j_t(\textbf{u}_t), P^{des}(f_t^j))\\
    + \lambda_{c1} \mathcal{L}_{dist}(P({d_{1}}^j_t(\textbf{u}_t), P^{des}({d_{1}}_t^j))\\ 
    + \lambda_{c2} \mathcal{L}_{dist}(P({d_{2}}^j_t(\textbf{u}_t), P^{des}({d_{2}}_t^j)),\\ \textbf{u}_t \in \mathcal{C}
    \end{aligned}
\end{equation}

\noindent where, ${\lambda}$ is the weight for $\mathcal{L}_{dist}$, $\mathcal{L}_{dist}$ is a metric of similarity between the two given distributions, $P(f^j_t)$ and $P^{des}(f_t^{j})$. $P^{des}(f_t^{j})$ is a desired distribution which avoids the collision with $\eta \to 1$. Here, we fold the chance constraint into the main cost function and manifest a surrogate distribution matching cost \cite{pvo-ral2020}, \cite{gopalakrishnan2018tcst}. It should be noted that we do not have access the parametric form of the distributions of $P(f_t^j)$ and $P(f_t^{j})$ but we assume access to the samples of the same. We elaborate on the desired distribution in section \ref{sec:desired-distribution}. We describe the method to sample the feasible set of controls $\textbf{u}_t \in \mathcal{C}$ in section \ref{sec:sampling-controls}.

\subsection{Desired Distribution}
\label{sec:desired-distribution}
\begin{figure}
    \includegraphics[width=.49\textwidth]{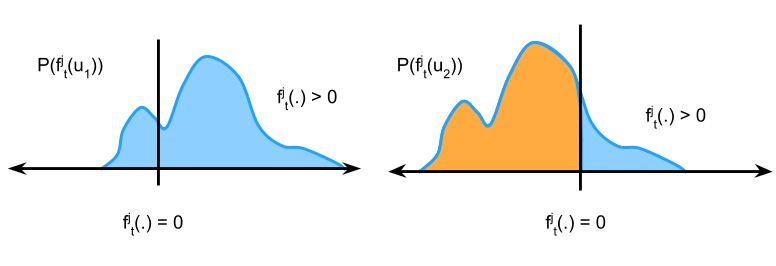}
    \caption{\small The shape of the distribution can be manipulated by $\textbf{u}$. An appropriate shape is one where most of the probability distributions  mass lies to the left of $f(.)=0$. \normalsize}
    \label{distribution-shift}
\end{figure}

A control input that can shape the distribution of constraints to have an appropriate shape, e.g, appropriate mass to the left of zero as shown in the Figure \ref{distribution-shift} is a solution of the chance constrained optimization. The solution distribution should be as similar as possible to the desired distribution in aspects such as the similarity around the tail of the distribution. To put it more simply, the desired distribution is an estimate of how the solution distribution might look like. Thus, given a feasible chance-constrained optimization there always exists a desired distribution. However, the essential question one should ask is whether we can always construct such a desired distribution. The construction of the desired distribution is equivalent to solving the following deterministic optimization problem for $\textbf{u}_{t}^{nom}$.

\small
\begin{subequations}
\begin{align}
\textbf{u}_{t}^{nom} = \argmin_{\textbf{u}_t} J(\textbf{u}_{t})\label{cost_scenario}\\
f_t^j(\textbf{u}_{t}^{nom})\leq 0,\\
{d_{1}}^j_t(\textbf{u}_{t}^{nom})\leq 0,\\
{d_{2}}^j_t(\textbf{u}_{t}^{nom})\geq 0,\\
\forall \  \textbf{u}_{t}\in \mathcal{C}\label{feasible_scenario},
\end{align}
\end{subequations}
\normalsize

The different $f_t^j(\textbf{u}_{t}^{nom}, \cdot )$ computed for  $\textbf{u}_{t}^{nom}$ obtained from (\ref{cost_scenario})-(\ref{feasible_scenario}) are precisely the samples of the desired distribution $P^{des}_{f_t^j}$. A desired distribution can be constructed as long as we can solve (\ref{cost_scenario})-(\ref{feasible_scenario}). {In other words, the construction guarantee of the desired distribution is tied to the existence guarantee of solution of (\ref{cost_scenario})-(\ref{feasible_scenario})}. Now, since $f_t^j(.)$ represents the velocity obstacle function, (\ref{cost_scenario})-(\ref{feasible_scenario}) represents a non-convex optimization problem for which absolute solution guarantees are intractable to obtain. Thus, instead we focus on the conditions under which a solution to (\ref{cost_scenario})-(\ref{feasible_scenario}) is likely to be obtained.

\noindent \textbf{Assumption 1:} A $\textbf{u}_{t}^{nom}$ exists such that $f_t^j(\textbf{u}_{t}^{nom}  )\leq 0$, ${D_{1}}^j_t(\textbf{u}_{t}^{nom})\leq 0$, and ${D_{2}}^j_t(\textbf{u}_{t}^{nom})\geq 0$.

\textbf{Assumption 1} essentially means that a collision avoidance control can be obtained if we disregard the uncertainty and just consider the mean. This is a fair assumption since velocity obstacle is indeed known to produce collision-free motion in the deterministic case with a very high success rate \cite{manocha-rvo}. Now, the solution computed with respect to the mean would always result in some non-zero probability. For example, if the uncertainty is Gaussian, the control input computed based on the mean of uncertainty will lead to $50\%$ probability of collision avoidance. This, in turn, implies that there would always be samples around the mean that will be collision-free with respect to the control input computed considering just the mean of the uncertainty. Thus, it is likely that if we choose samples $^r\textbf{x}_t$, $(^o\textbf{x}^j_t, ^o\dot{\textbf{x}}^j_t)$ within a ball of radius $\epsilon$ of the mean uncertainty, we would obtain a solution for (\ref{cost_scenario})-(\ref{feasible_scenario}) and hence can  construct  the desired distribution. 

The desired distribution can be constructed as long as the solution to the deterministic counterpart of (\ref{eq:mpc}) can be obtained by solely the deterministic version of MPC without the surrogate cost. Further, the small perturbations around the mean state and control are considered and solved for. As long as a solution exists to the mean MPC and perturbations around the mean, a desired distribution can be obtained. The distribution of $f^j_t$, ${d_{1}}^j_t$ and ${d_{2}}^j_t$ are all functions of the state of the robot and obstacle as well as the control input to the robot, which are all, in turn, random variables.

\subsection{Sampling feasible controls}
\label{sec:sampling-controls}
We solve the optimization problem in (\ref{eq:mpc}) by sampling a set of feasible control commands and choosing the best control from the sampled set. The sampling approach makes solving (\ref{eq:mpc}) independent of the motion model of the robot. For simplicity, we sample a set of feasible controls using a differential drive motion model. Algorithm \ref{alg:control-sampling} describes how the feasible sample set $\mathcal{C}$ is populated.


\begin{algorithm}
	\caption{Sampling controls using the unicycle model}
	\begin{algorithmic}[1]
	    \label{alg:control-sampling}
	    \STATE $N_v \gets$ \# steps in the magnitude of velocity
	    \STATE $N_{\omega} \gets$ \# steps in the magnitude of angular velocity
	    \STATE $(v_{lb}, v_{ub}) \gets$ Bounds on magnitude of velocity
	    \STATE $(\omega_{lb}, \omega_{ub}) \gets$ Bounds on magnitude of angular velocity
	    \STATE $V(i) \gets v_{lb} + i*\frac{v_{ub}-v_{lb}}{N_v}, \forall i \in \{1, 2, 3, \dots, N_v\}$
	    \STATE $\omega(j) \gets \omega_{lb} + i*\frac{\omega_{ub}-\omega_{lb}}{N_{\omega}}, \forall j \in \{1, 2, 3, \dots, N_{\omega}\}$
		\FOR {Every time step}
		    \FOR {$i \in {1, 2, 3, \dots, N_v}$ }
		        \FOR {$j \in {1, 2, 3, \dots, N_{\omega}}$}
		            \STATE $^{i,j}\textbf{u}_t \gets \begin{bmatrix} V(i) \cos(\theta_t + \Omega(j) \Delta t) \\ V(i) \sin(\theta_t + \Omega(j) \Delta t) \end{bmatrix}$
		        \ENDFOR
		    \ENDFOR
		\STATE $\mathcal{C} \gets \{x \mid x =\ ^{i,j}\textbf{u}_t, \forall i, j \}$
		\ENDFOR
	\end{algorithmic} 
\end{algorithm} 
Essentially, we take steps between the bounds on the magnitude of angular velocities and the bounds of the magnitude of linear velocities to get $N_v$ samples of linear velocities and $N_w$ samples of angular velocities. All combinations of the sampled linear and angular velocities are used to generate the set of feasible controls $\mathcal{C}$. The number of steps is chosen to ensure adequate coverage of the sample space of velocities that the ego vehicle can take.  

\subsection{Solving Collision Avoidance Chance Constraints}
\label{sec:solving-chance-constraints}

\subsubsection{Distribution matching}
A probability distribution is defined by its moments. From \cite{moments-determine-tail}, for given random variables x and y, the distributions of x and y become more similar as the higher-order moments become similar.

\begin{equation}
    \norm{P(\text{x}) - P(\text{y})} \leq B(d), B(d) \to 0, d \to \infty
\end{equation}

where, $d$ refers to the order up to which the moments of $P(\text{x})$ and $P(\text{y})$ are similar. To the best of our knowledge, there is no method to deduce the exact mapping between the similarity of the distribution and order of moment to match starting from the first moment. Taking the concepts from \cite{kme-nips}, \cite{pvo-ral2020} formulates a workaround using the concept of embedding distributions in Reproducing Kernel Hilbert Space (RKHS) and Maximum Mean Discrepancy (MMD) distance.

\begin{subequations}
    \label{eq:mmd-distance}
    \begin{equation}
        \mathcal{L}_{dist}(x, y) = K_{xx} - 2K_{xy} + K_{yy}
    \end{equation}
    \begin{align}
        K_{xx} = {\alpha}^T_{1 \times n} k_d(x, x)_{n \times n} {\alpha}_{n \times 1} \\
        K_{xy} = {\alpha}^T_{1 \times n} k_d(x, y)_{n \times n} {\beta}{n \times 1} \\
        K_{yy} = {\beta}^T_{1 \times n} k_d(y, y)_{n \times n} {\beta}_{n \times 1}
    \end{align}
    \begin{align}
        k_d(x, x) = (a {L}_{n \times 1} {L}^T_{1 \times n} + l)^d \\
        k_d(x, y) = (a {L}_{n \times 1} {M}^T_{1 \times n} + l)^d \\
        k_d(y, y) = (a {M}_{n \times 1} {M}^T_{1 \times n} + l)^d 
    \end{align}
\end{subequations}

where, $L$ and $M$ are $n$ number of samples from respective distributions of the random variables x and y. $k_d(\cdot)$ represents a polynomial kernel of degree $d$. $\alpha$ and $\beta$ are the weight vectors corresponding to $W_x$ and $W_y$. For $i.i.d$ samples, $\alpha$ and $\beta$ take the values given in (\ref{eq:iid-weights}).

\begin{equation}
    \label{eq:iid-weights}
    \alpha = \beta = \frac{1}{n} \textbf{1}_{n \times 1}
\end{equation}

\subsubsection{Collision Avoidance}
Algorithm \ref{alg:collision-avoidance} describes the procedure to obtain an optimal control, $\textbf{u}^{opt}_t$, that avoids the obstacles under uncertainty at time $t$.

\begin{algorithm}
	\caption{Collision avoidance at time $t$}
	\begin{algorithmic}[1]
	    \label{alg:collision-avoidance}
	    \STATE $^r\textbf{x}_t \gets$ Samples of the position of the robot at time $t$
	    \STATE $(^o\textbf{x}^j_t, ^o\dot{\textbf{x}}^j_t) \gets$ Samples of (Position, Velocity) of the $j^{th}$ obstacle at time $t$
		\FOR {Every $\textbf{u}_t \in \mathcal{C}$}
		    \STATE Evaluate the value of $J(\textbf{u}_t)$ using (\ref{eq:mpc-cost})
		    \STATE $\mathcal{F} \gets \{x | x = f^j_t(\textbf{u}_t), \forall (^r\textbf{x}_t, ^o\textbf{x}^j_t,
		    ^o\dot{\textbf{x}}^j_t, \delta_t)\}$
		    \STATE $\mathcal{D}_{1} \gets \{x | x = {d_{1}}^j_t(\textbf{u}_t), \forall (^r\textbf{x}_t, ^o\textbf{x}^j_t,
		    ^o\dot{\textbf{x}}^j_t, \delta_t)\}$
		    \STATE $\mathcal{D}_{2} \gets \{x | x = {d_{2}}^j_t(\textbf{u}_t), \forall (^r\textbf{x}_t, ^o\textbf{x}^j_t,
		    ^o\dot{\textbf{x}}^j_t, \delta_t)\}$
		    \STATE $\mathcal{F}^{des} \gets \{f^j_t(\textbf{u}^{nom}_t) | f^j_t(\cdot) \leq 0\}$
		    \STATE $\mathcal{D}_{1}^{des} \gets \{{d_{1}}^j_t(\textbf{u}^{nom}_t) | {d_{1}}^j_t(\cdot) \leq 0\}$
		    \STATE $\mathcal{D}_{2}^{des} \gets \{{d_{2}}^j_t(\textbf{u}^{nom}_t) | {d_{2}}^j_t(\cdot) \geq 0\}$
		    \STATE Evaluate the value of $L_{dist}(\mathcal{F}, \mathcal{F}^{des})$ using (\ref{eq:mmd-distance})
		    \STATE $q(\textbf{u}_t) \gets J(\textbf{u}_t) + \lambda L_{dist}(\mathcal{F}, \mathcal{F}^{des}) + \lambda_{1} L_{dist}(\mathcal{D}_{1}, \mathcal{D}_{1}^{des}) + \lambda_{2} L_{dist}(\mathcal{D}_{2}, \mathcal{D}_{2}^{des})$
		\ENDFOR
		\STATE $\mathcal{Q} \gets \{x | x = q(\textbf{u}_t) \forall \textbf{u}_t \in \mathcal{C}\}$
		\STATE $\textbf{u}_t^{opt} \gets \argmin_{\textbf{u}_t} Q$
		\RETURN $\textbf{u}_t^{opt}$
	\end{algorithmic} 
\end{algorithm} 


Essentially, we calculate the distributions $\mathcal{F}$, $\mathcal{D}_{1}$ and $\mathcal{D}_{2}$ for each sample control in the feasible sample set $\mathcal{C}$ for all noise samples. Then the desired distributions $\mathcal{F}^{des}$, $\mathcal{D}_{1}^{des}$, and $\mathcal{D}_{2}^{des}$ are calculated. Finally we calculate the cost defined in (\ref{eq:mpc-rkhs}) as $Q$ and find the optimal control as $\textbf{u}_t^{opt}$ which yields the minimum cost $q(\textbf{u}_t)$.

%% file: chapters/Sim1.tex
\section{Results and Discussion}
\label{Results}

All simulations were carried out on a commodity laptop with MATLAB as the simulation engine. The robot's state was evolved according to unicycle kinematics. Non-Gaussian and at times multi-model distributions were used to model the robot's state, velocity and control noise and the state and velocity noise of the dynamic obstacles and participants.More details about the noise distributions can be found here \href{https://robotics.iiit.ac.in/publications/2020/non-holonomic-collision-avoidance-hilbert-space.html}{https://robotics.iiit.ac.in/publications/2020/non-holonomic-collision-avoidance-hilbert-space.html}. Twenty-five samples for robot and obstacle each were used to model the uncertainty in the state, velocity and actuator controls where the samples were drawn from a Pearson distribution.

\subsection{In-feasibility of an optimization based approach}
\label{sec:opt_probs}
\begin{figure}
\begin{center}
    \includegraphics[width=.25\textwidth]{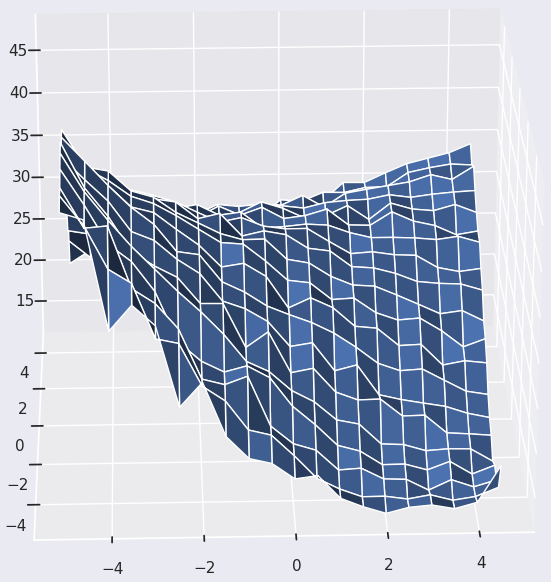}
\end{center}
    \caption{\small Notice the uneven cost surface with multiple kinks and valleys.\normalsize}
    \label{cost_surf}
\end{figure}

Figure \ref{cost_surf} shows that the cost function surface has many kinks and valleys where a gradient based optimisation technique may get stuck or cause delays in the time to converge or may even converge to a sub-optimal solution \ref{cost_surf}. Instead by sampling in the $(v,w)$ space we are able  to converge very close to an optimal solution and are able to maintain tight bounds on the frequency of control output which is very important for realtime control.

\subsection{Qualitative Results}
\begin{figure*}
    \begin{subfigure}{.99\textwidth}
        \includegraphics[width=.33\textwidth]{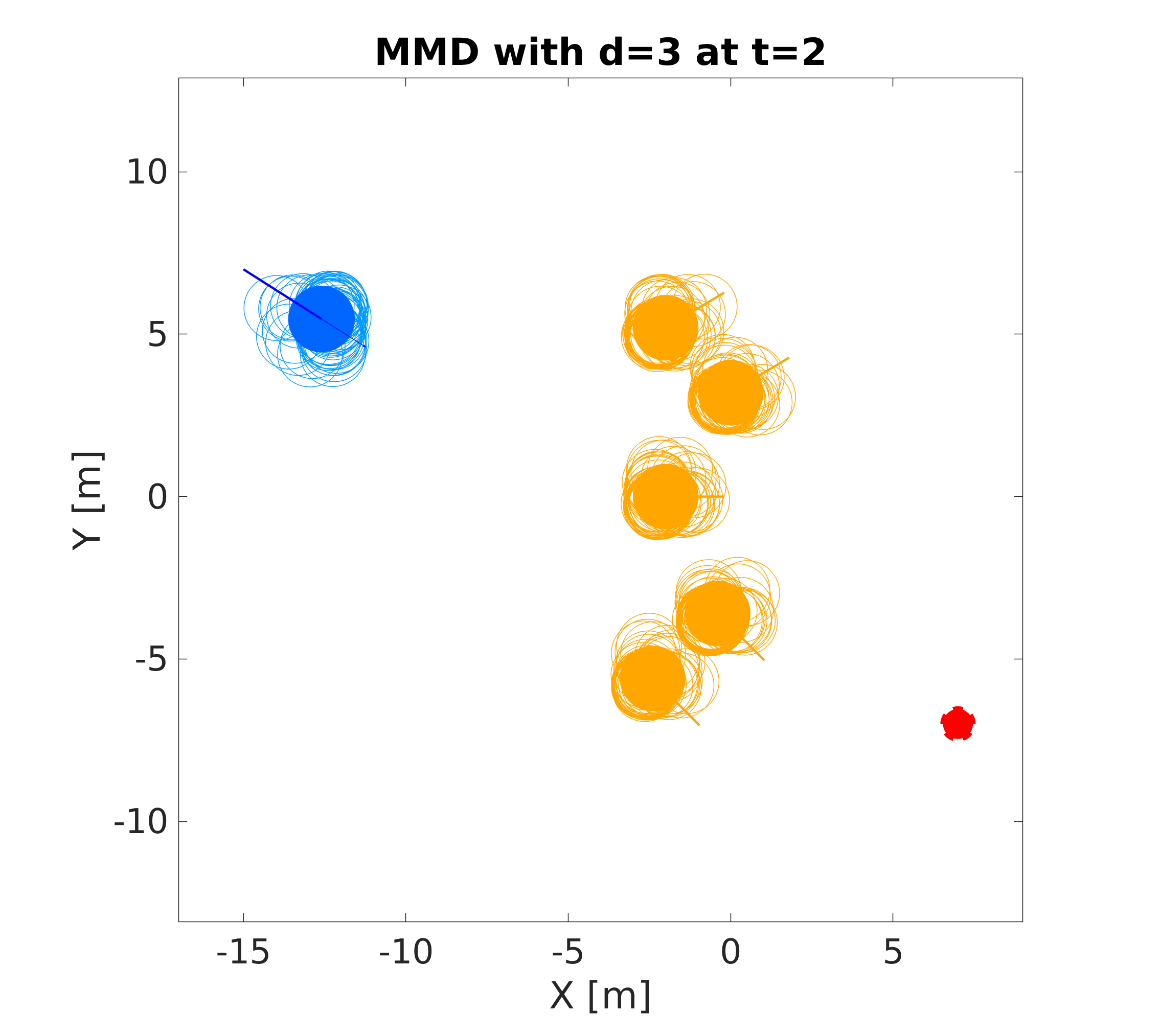}\hfill
        \includegraphics[width=.33\textwidth]{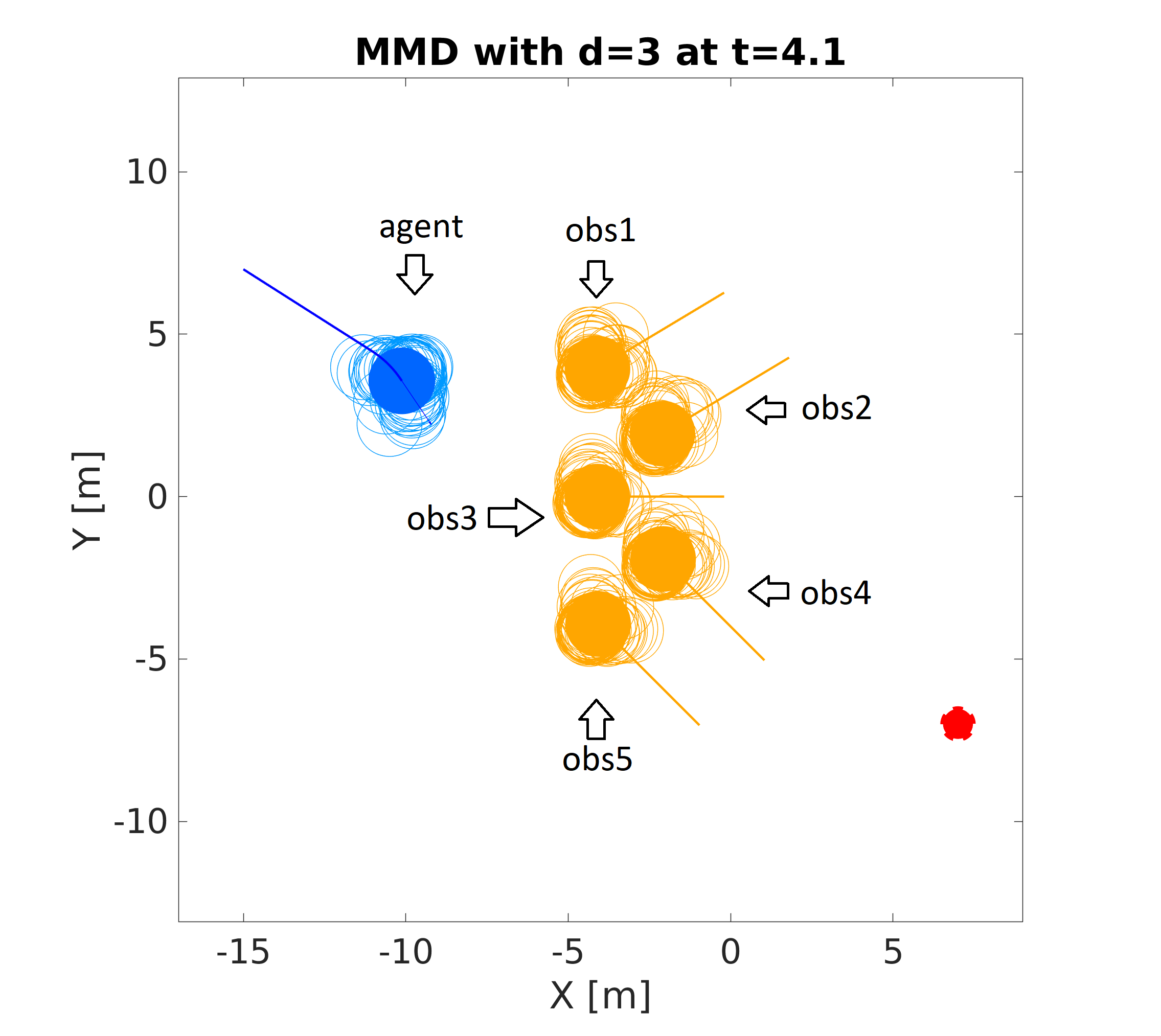}\hfill
        \includegraphics[width=.33\textwidth]{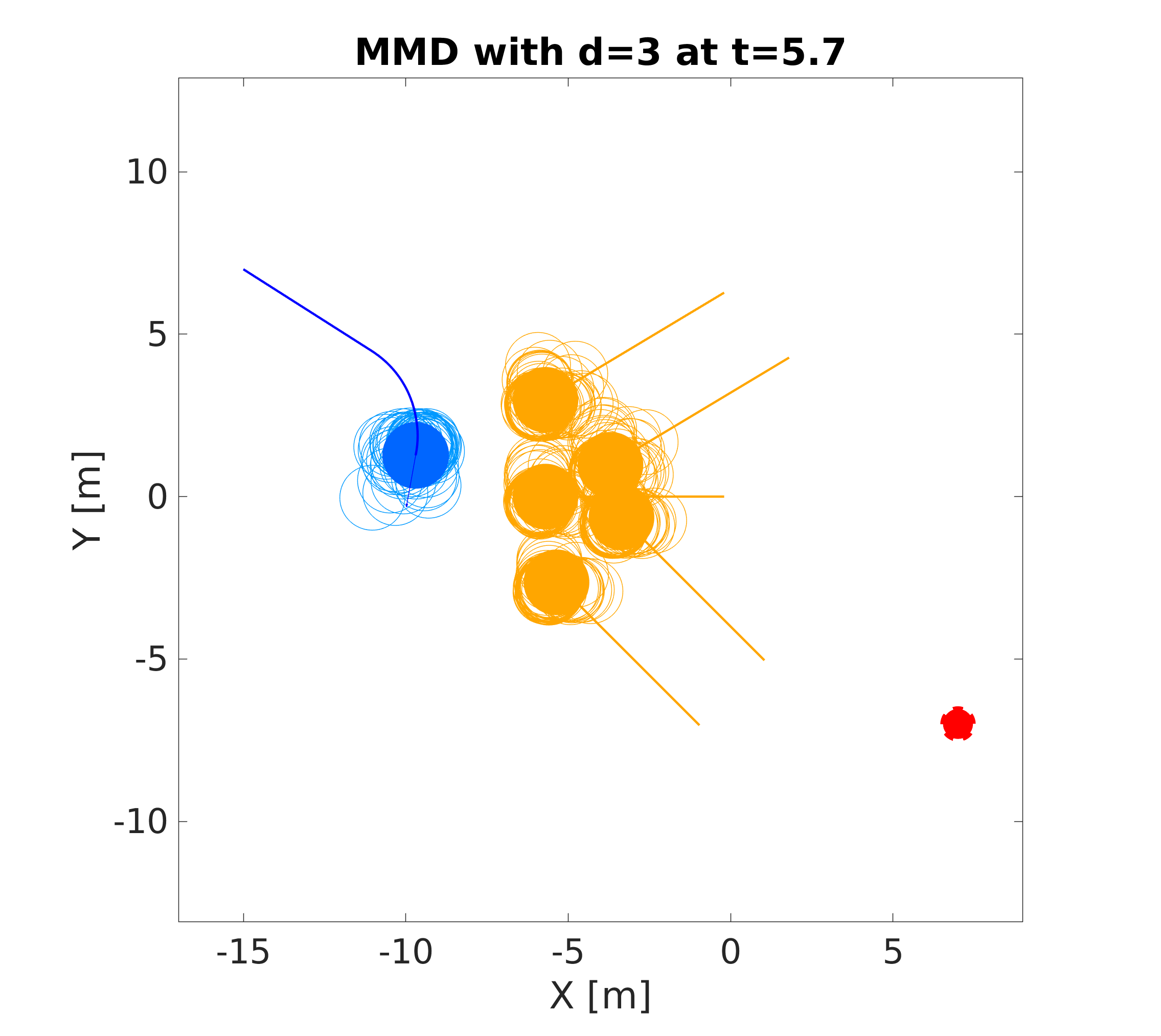}\hfill
        \includegraphics[width=.33\textwidth]{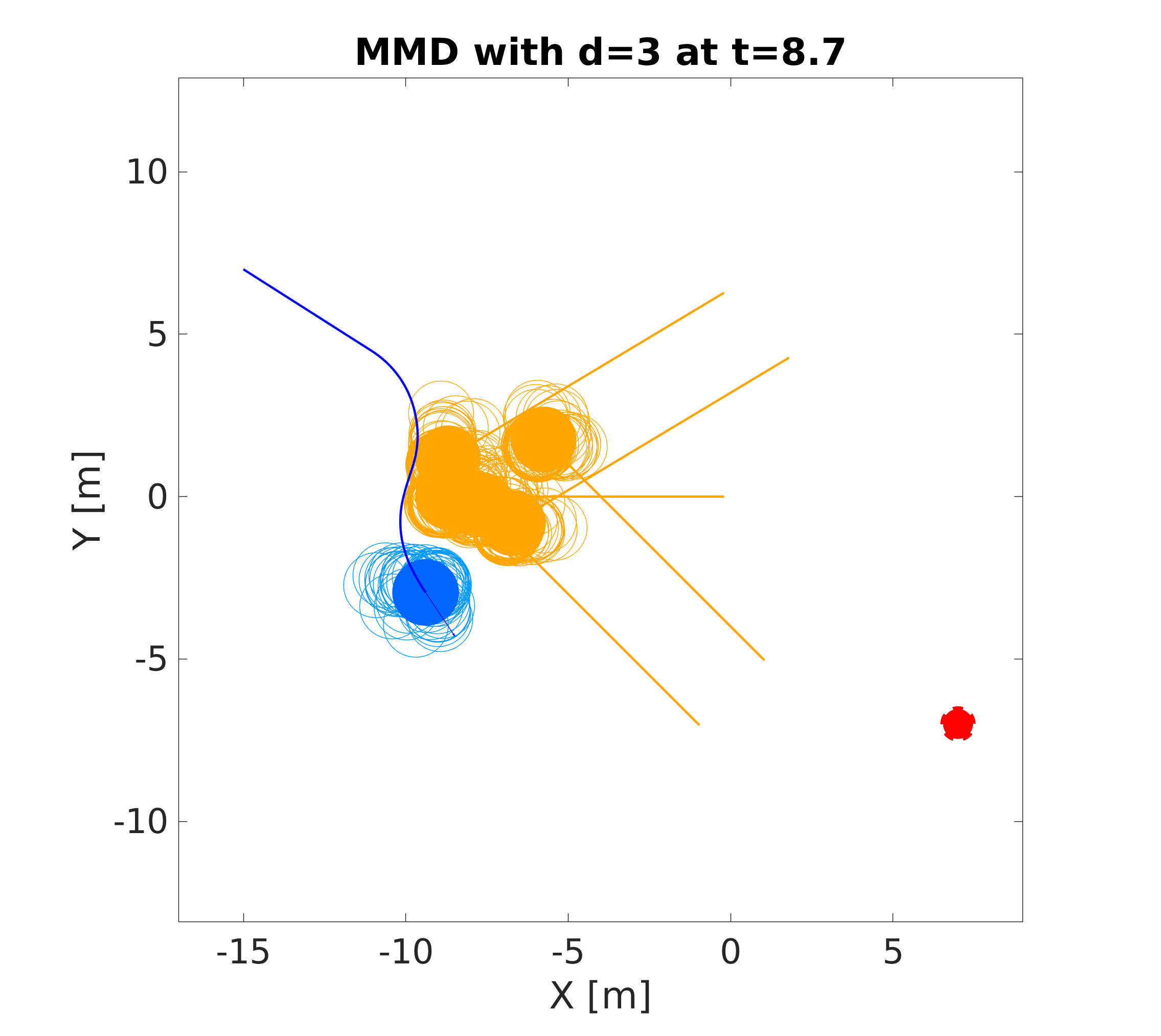}\hfill
        \includegraphics[width=.33\textwidth]{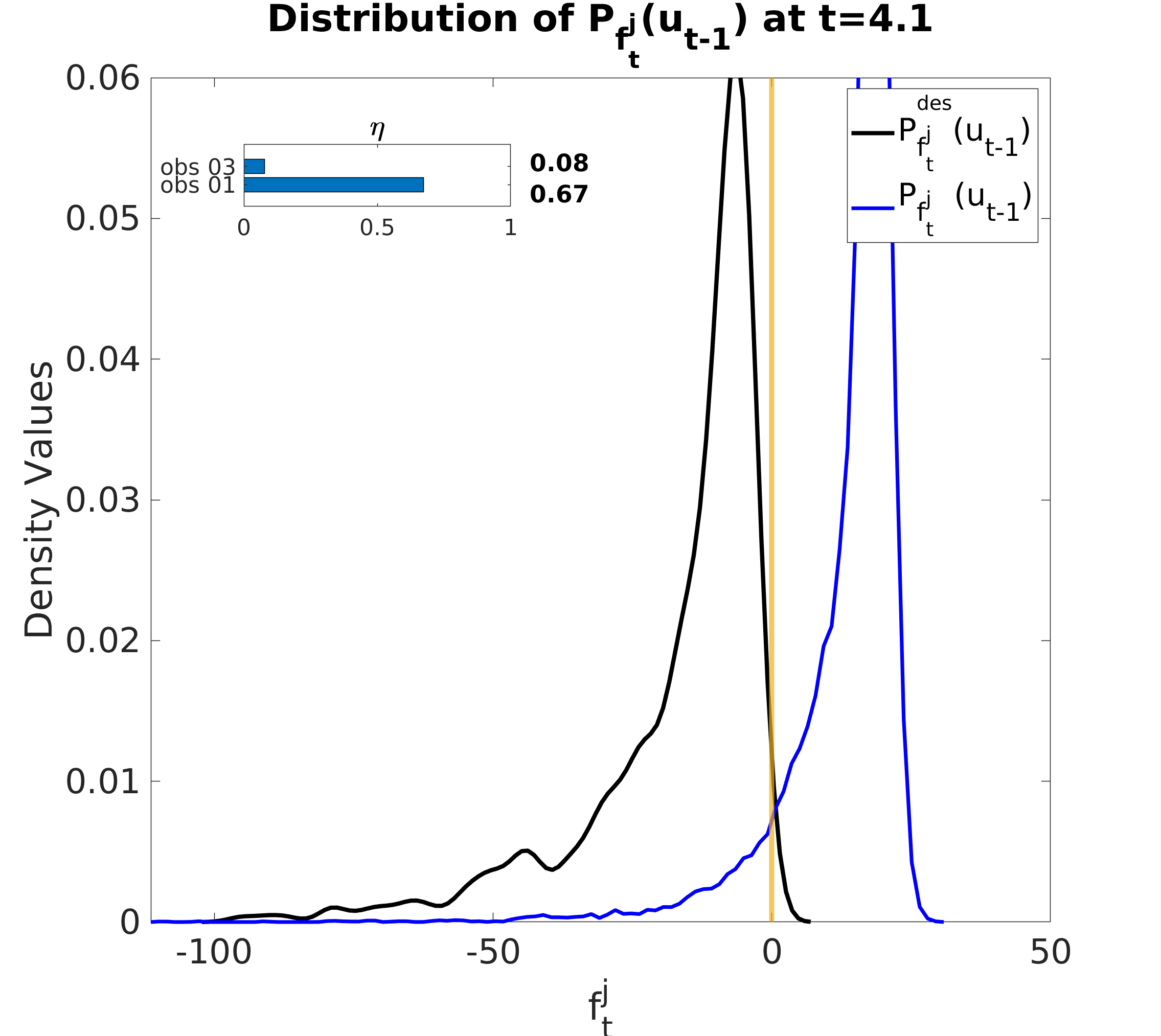}\hfill
        \includegraphics[width=.33\textwidth]{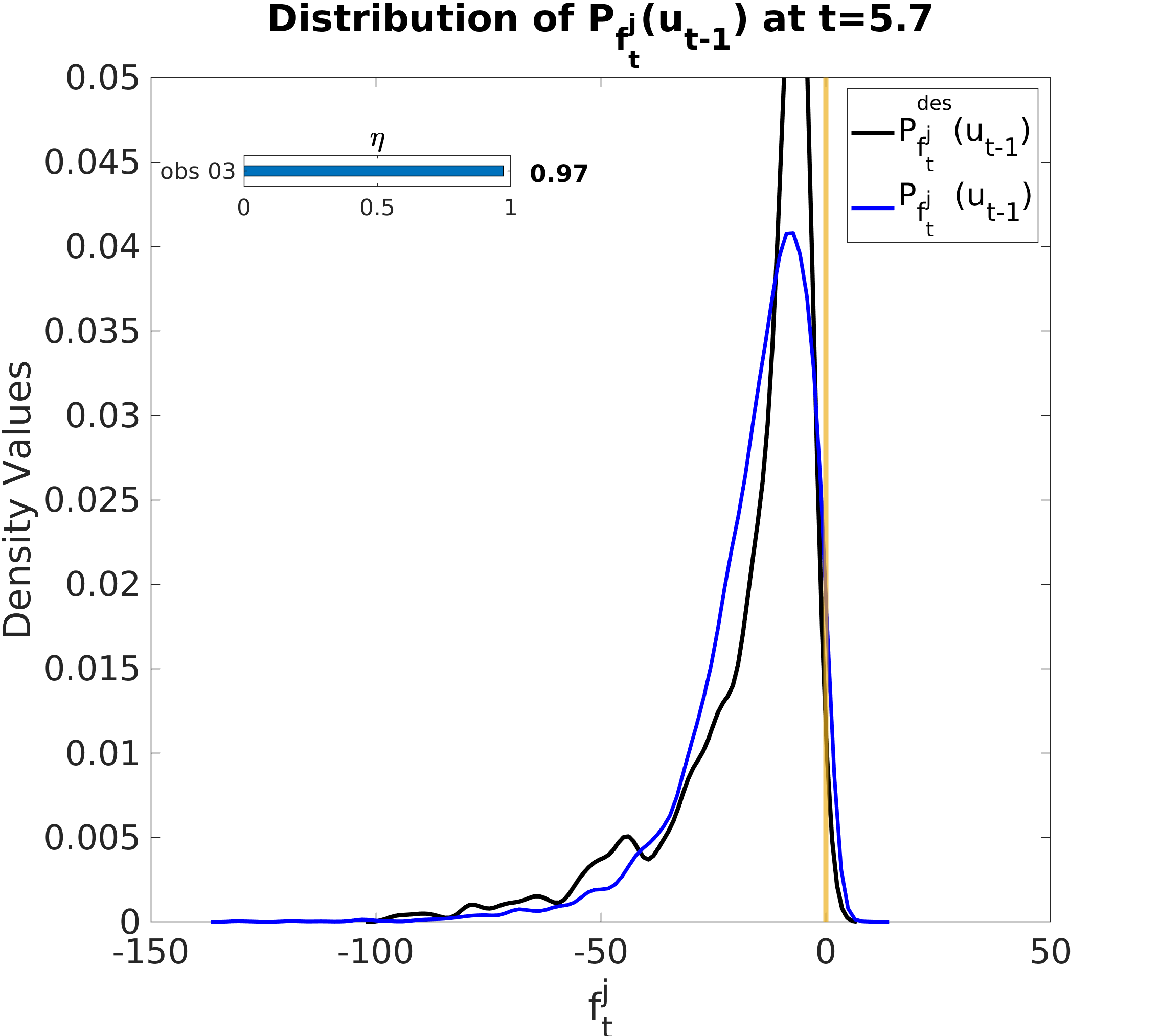}\hfill
        \caption{5 obstacles}
    \end{subfigure}
    \caption{\small Obstacle avoidance in the presence of 5 dynamic obstacles . For every snapshot in which we are executing an avoidance maneuver we show below it the desired distribution of collision cones in black and the current distribution in blue for the closest obstacle. \normalsize}
    \label{qual1}
\end{figure*}

\begin{figure*}
        \begin{subfigure}{.24\textwidth}
        \includegraphics[width=.9\textwidth]{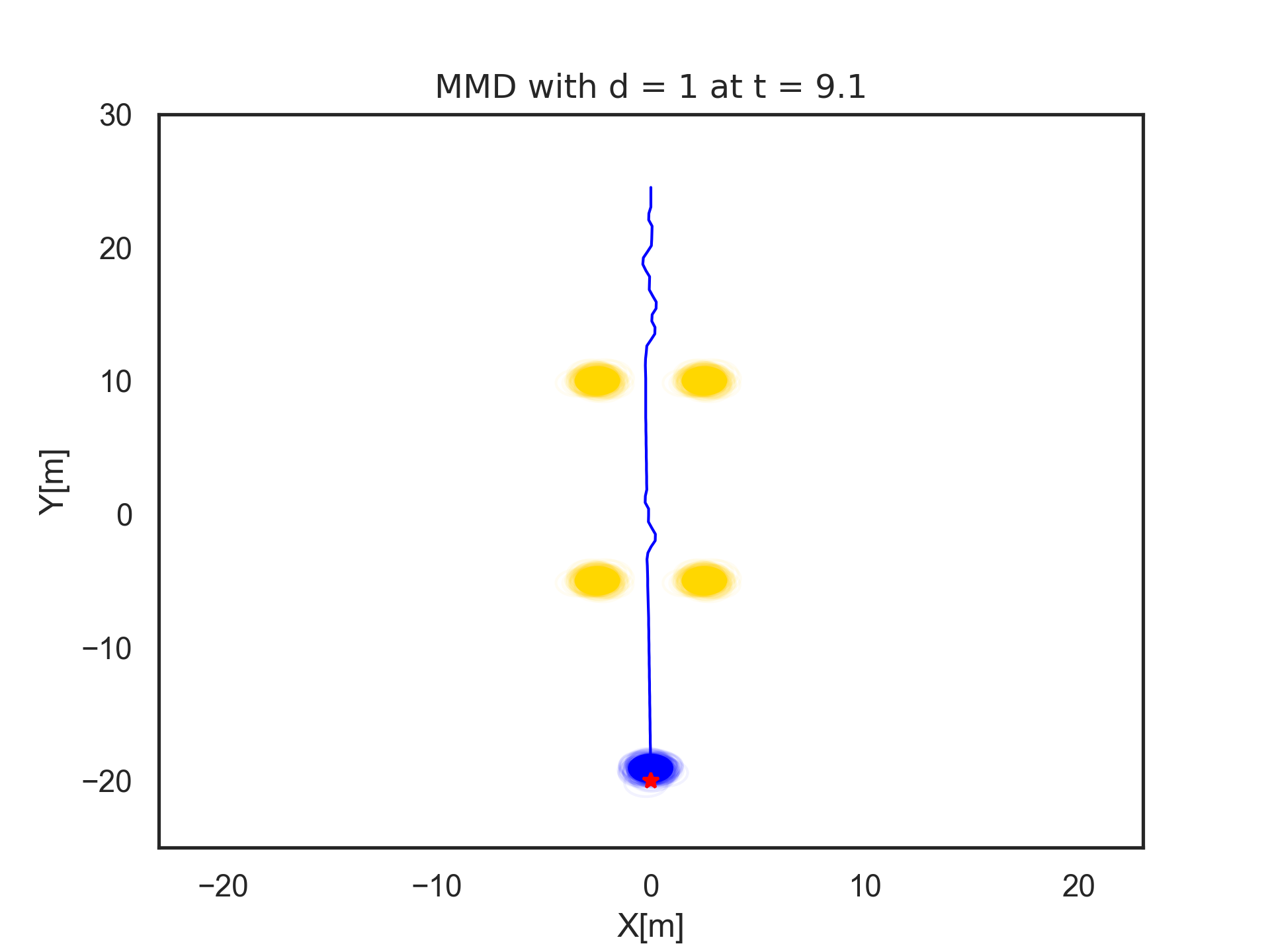}\hfill
        \caption{$d=1$}
        \label{static_d1}
        \end{subfigure}
        \begin{subfigure}{.24\textwidth}
        \includegraphics[width=.9\textwidth]{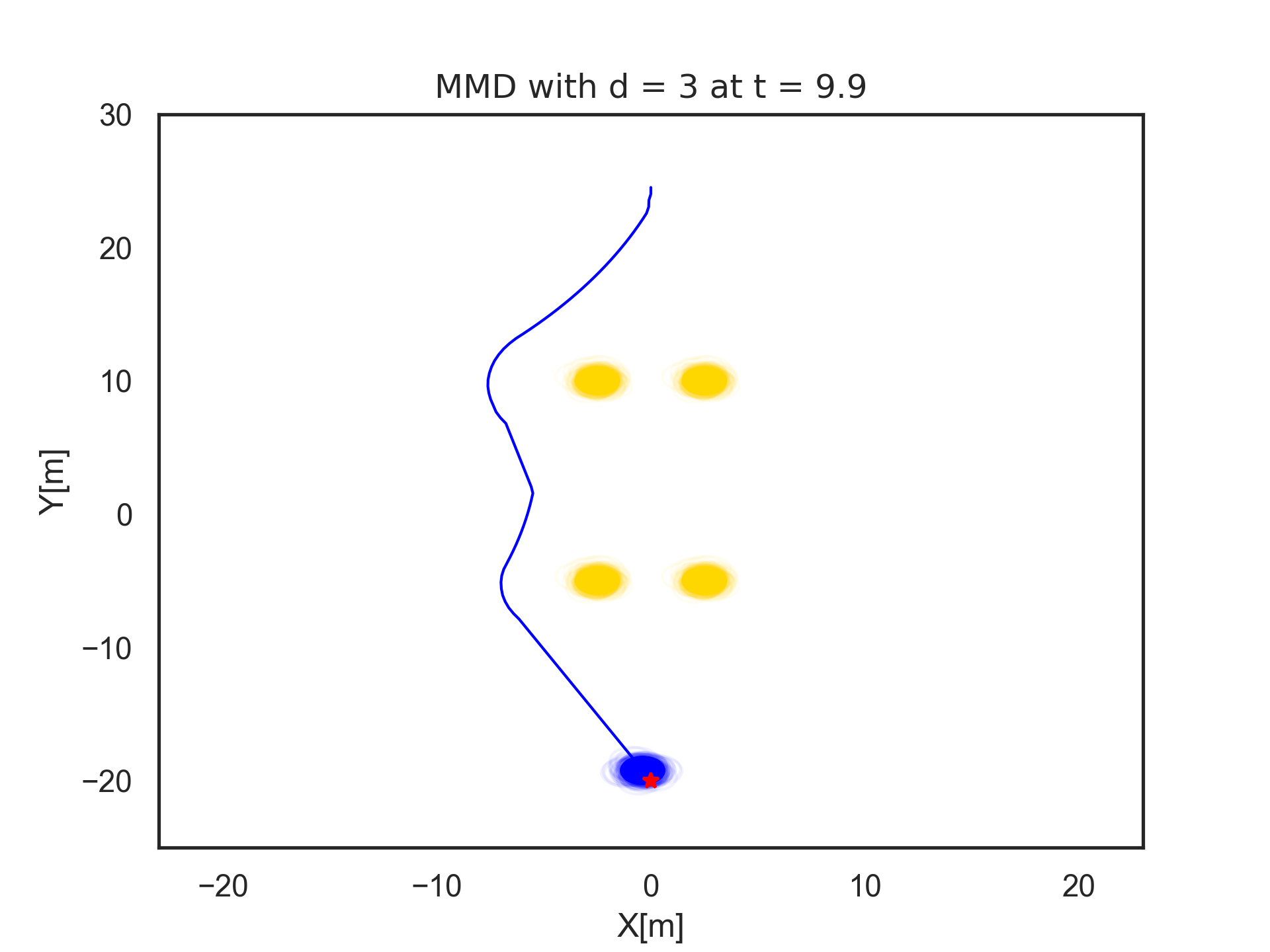}\hfill
        \caption{$d=3$}
        \label{static_d3}
        \end{subfigure}
        \begin{subfigure}{.24\textwidth}
        \includegraphics[width=.9\textwidth]{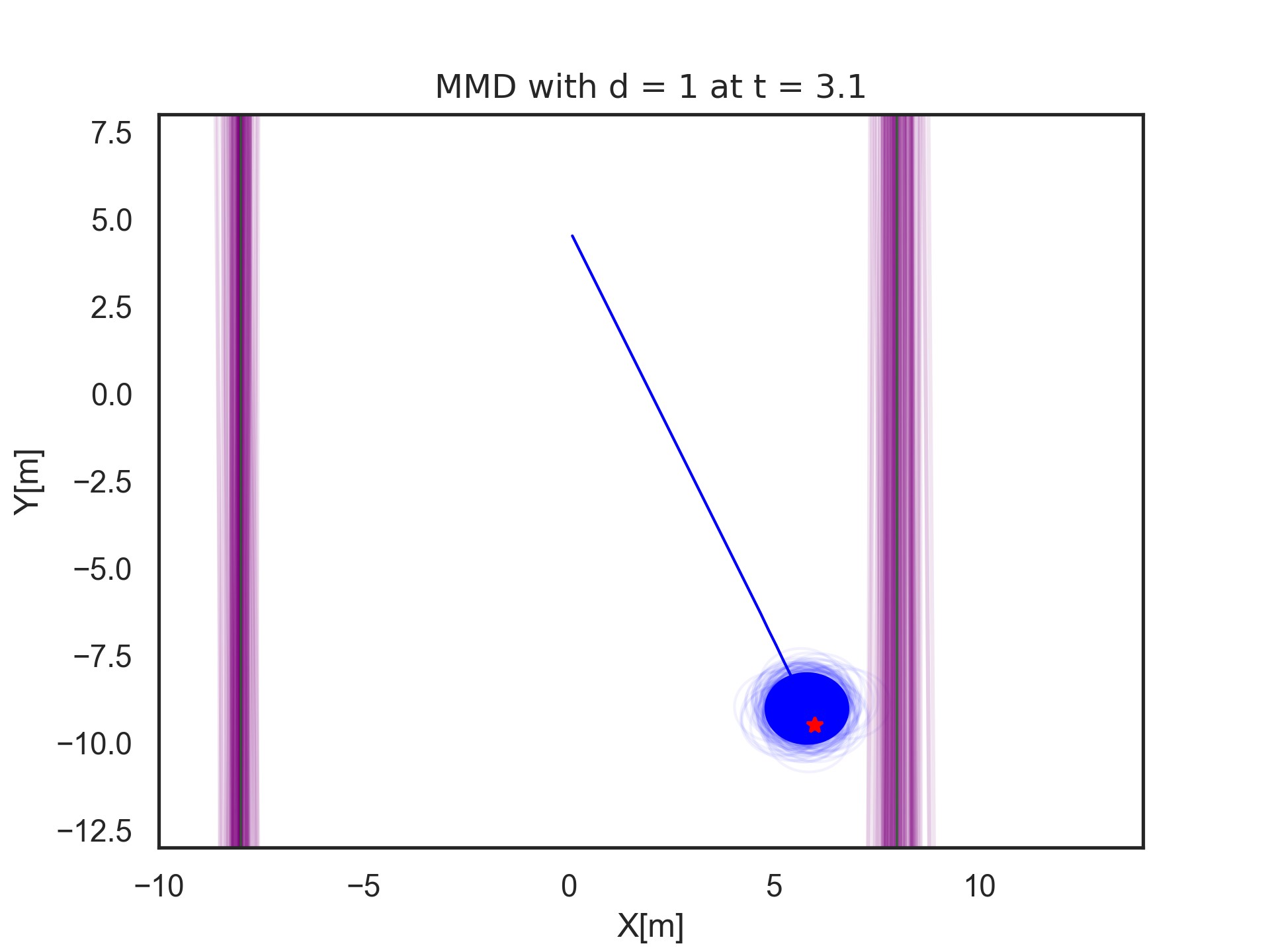}\hfill
        \caption{$d=1$}
        \label{lane_d1}
        \end{subfigure}
        \begin{subfigure}{.24\textwidth}
        \includegraphics[width=.9\textwidth]{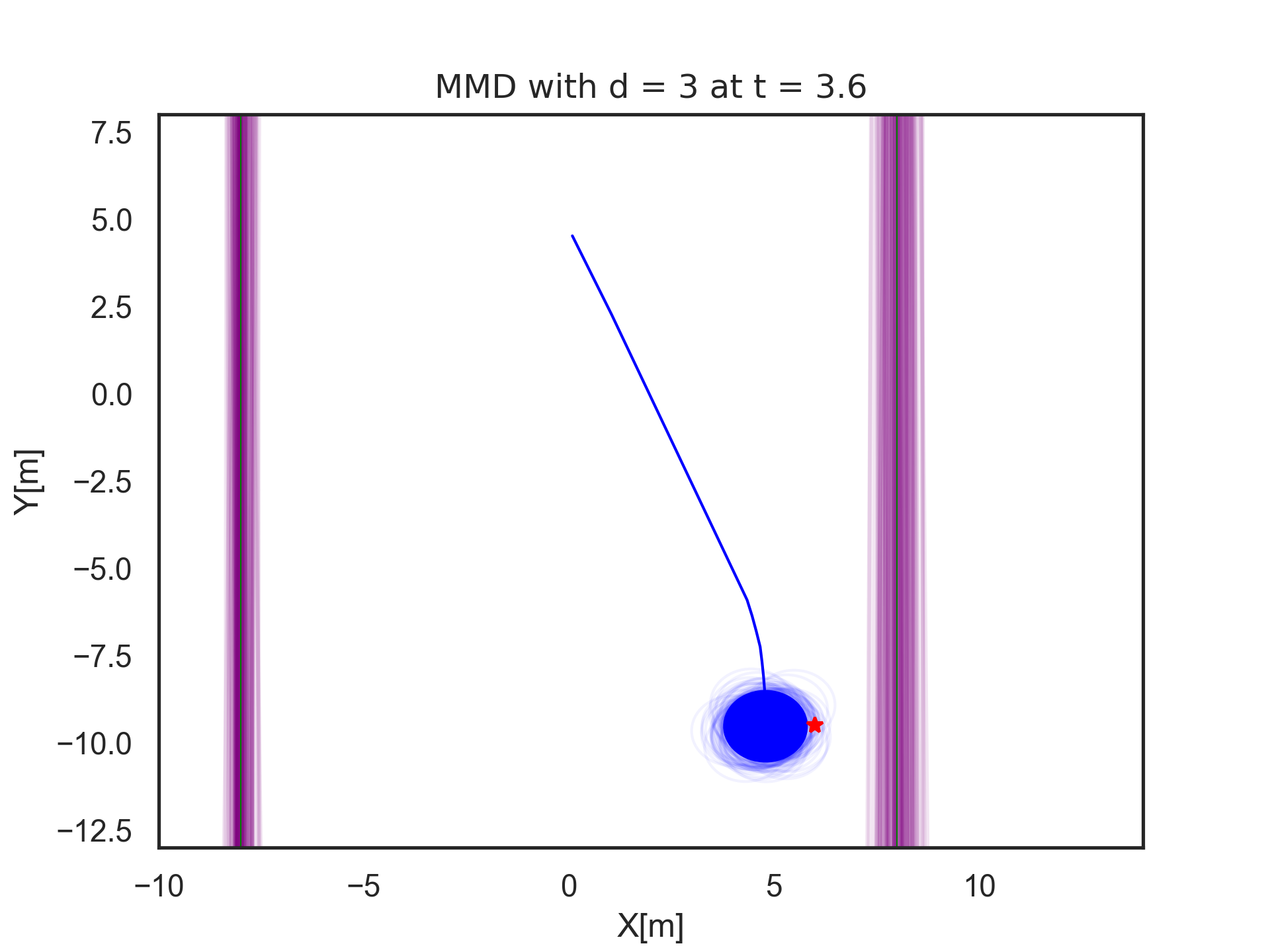}\hfill
        \caption{$d=3$}
        \label{lane_d3}
        \end{subfigure}
    \caption{\small \ref{static_d1} and \ref{static_d1} show the ego robot avoiding a static table and reaching a goal behind it. With $d=1$ we are able to go between the legs of the table. Whereas with higher values of $d$ we route around it. \ref{lane_d1} and \ref{lane_d1} show the ego vehicle prioritising to stay away from the lane boundaries with increasing $d$ while reaching the same goal.\normalsize}
    \label{static1}
\end{figure*}

Figure \ref{qual1} shows the various stages in an obstacle avoidance in the presence of five obstacles. For every snapshot where we are performing an avoidance manoeuvre we show below it the desired distribution of collision cones in black and the current distribution in blue for the closest obstacle. When a majority of samples of the robot are colliding with samples of the obstacles this distribution tends to be on the right side of the co-ordinate axis. Progressively as the obstacle gets avoided the collision cone distribution moves to the left of the $y$ axis and as can be seen in Figure \ref{qual1} coincides largely in its moments (such as the mean) with the desired distribution. Also shown in the top part of the distribution plots, is a horizontal bar whose length shows the number of samples of the obstacle distribution that are avoided. By showcasing the ability of the algorithm where many samples of the ego robot is able to avoid many samples of the obstacles we verify its efficiency. Figure \ref{static1} shows the how we can change the behaviour of the avoidance maneuver with changes in the tunable parameter $d$ in the presence of static obstacles and lanes/corridors. With $d=1$ we are able to pass into tight spaces while $d=3$ prioritises safer maneuvers and more avoidance. In these figures the obstacle is shown in a shade of orange, the robot in blue; the mean sample is shown dense in their respective colours whereas the other samples are shown by the circles whose interiors are unfilled. We also show the proposed method working in a real-time simulation in gazebo in the presence of lane/corridor constraints and stationary and dynamic obstacles at \href{https://robotics.iiit.ac.in/publications/2020/non-holonomic-collision-avoidance-hilbert-space.html}{https://robotics.iiit.ac.in/publications/2020/non-holonomic-collision-avoidance-hilbert-space.html}.

\subsection{Ablation Studies}
\label{sec:Ablation}
\begin{figure*}
    \begin{subfigure}{.33\textwidth}
        \includegraphics[width=0.75\linewidth]{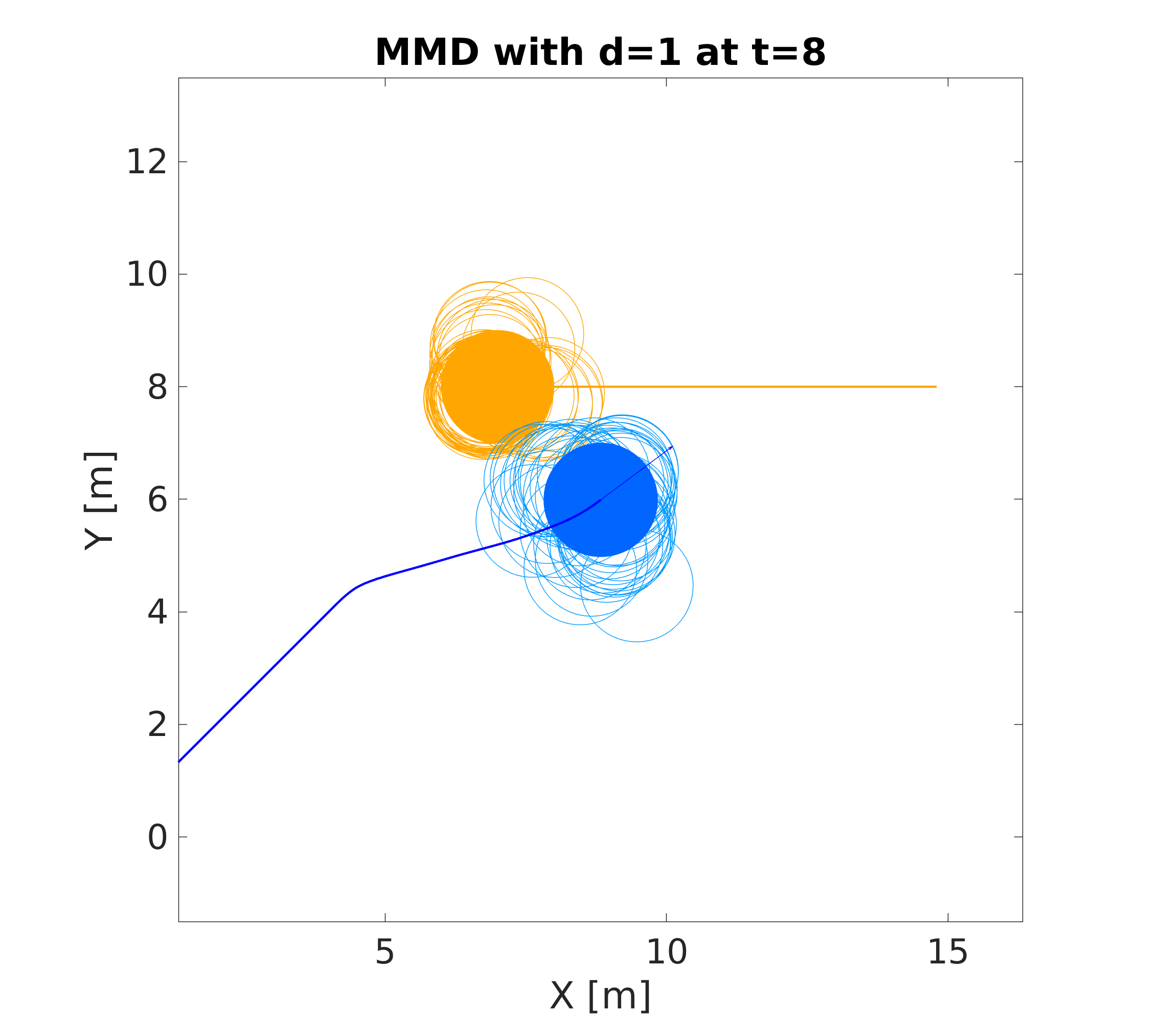}
        \caption{d = 1}
        \label{Abl1a}
    \end{subfigure}
    \begin{subfigure}{.33\textwidth}
        \includegraphics[width=0.75\linewidth]{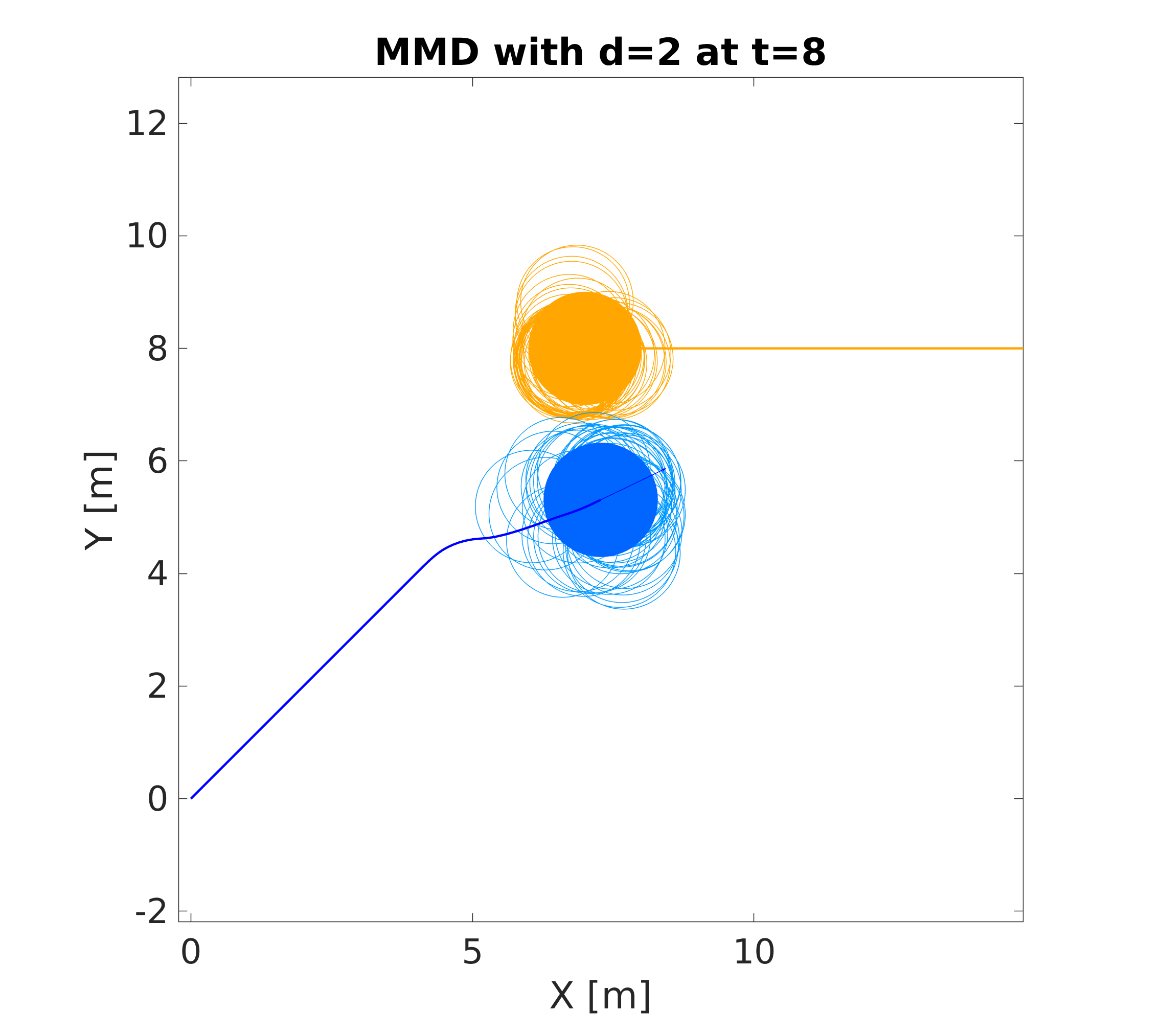}
        \caption{d = 2}
        \label{Abl1b}
    \end{subfigure}    
        \begin{subfigure}{.33\textwidth}
        \includegraphics[width=0.75\linewidth]{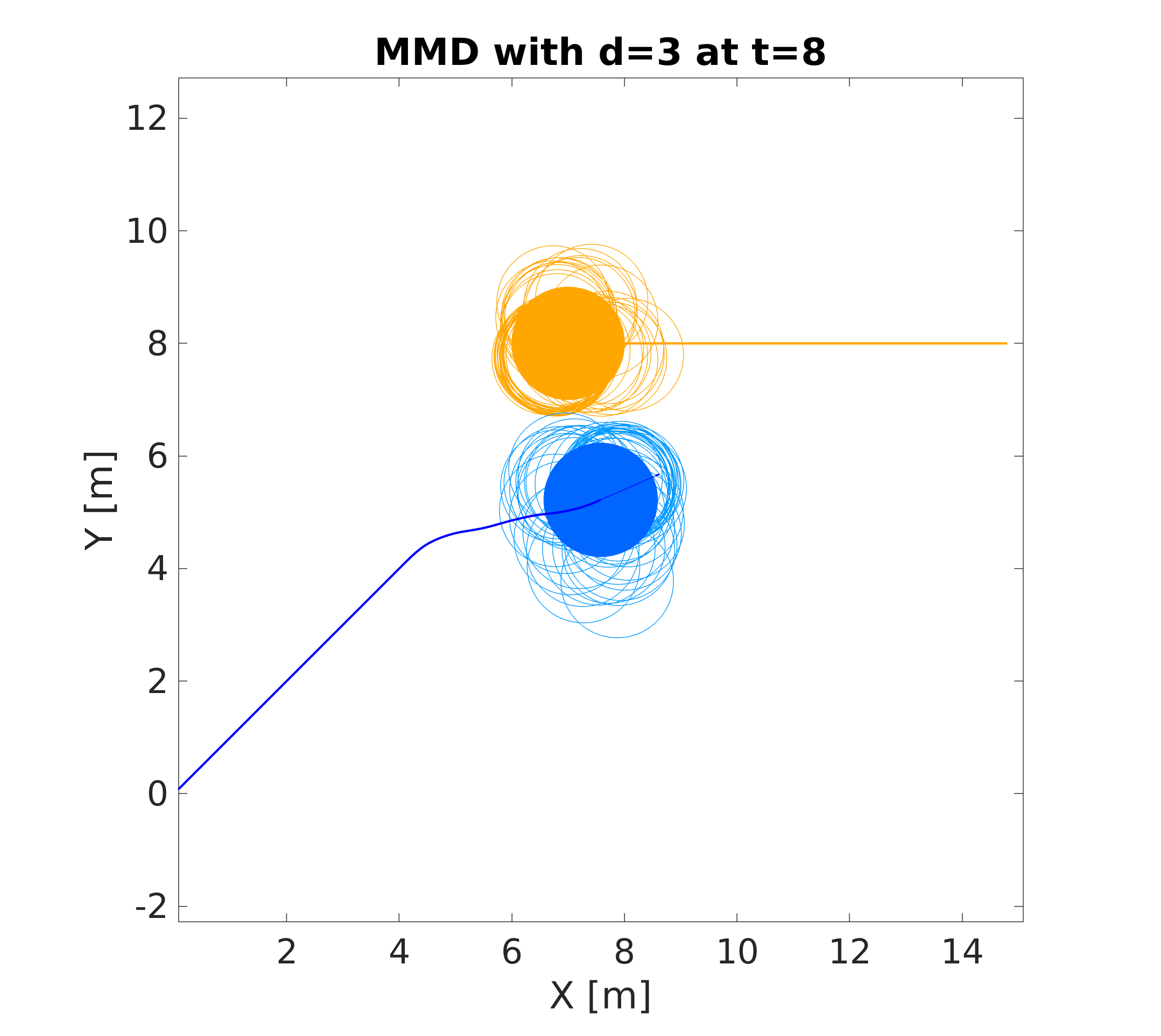}
        \caption{d = 3}
        \label{Abl1c}
    \end{subfigure}
    \begin{subfigure}{.33\textwidth}
        \includegraphics[width=0.75\linewidth]{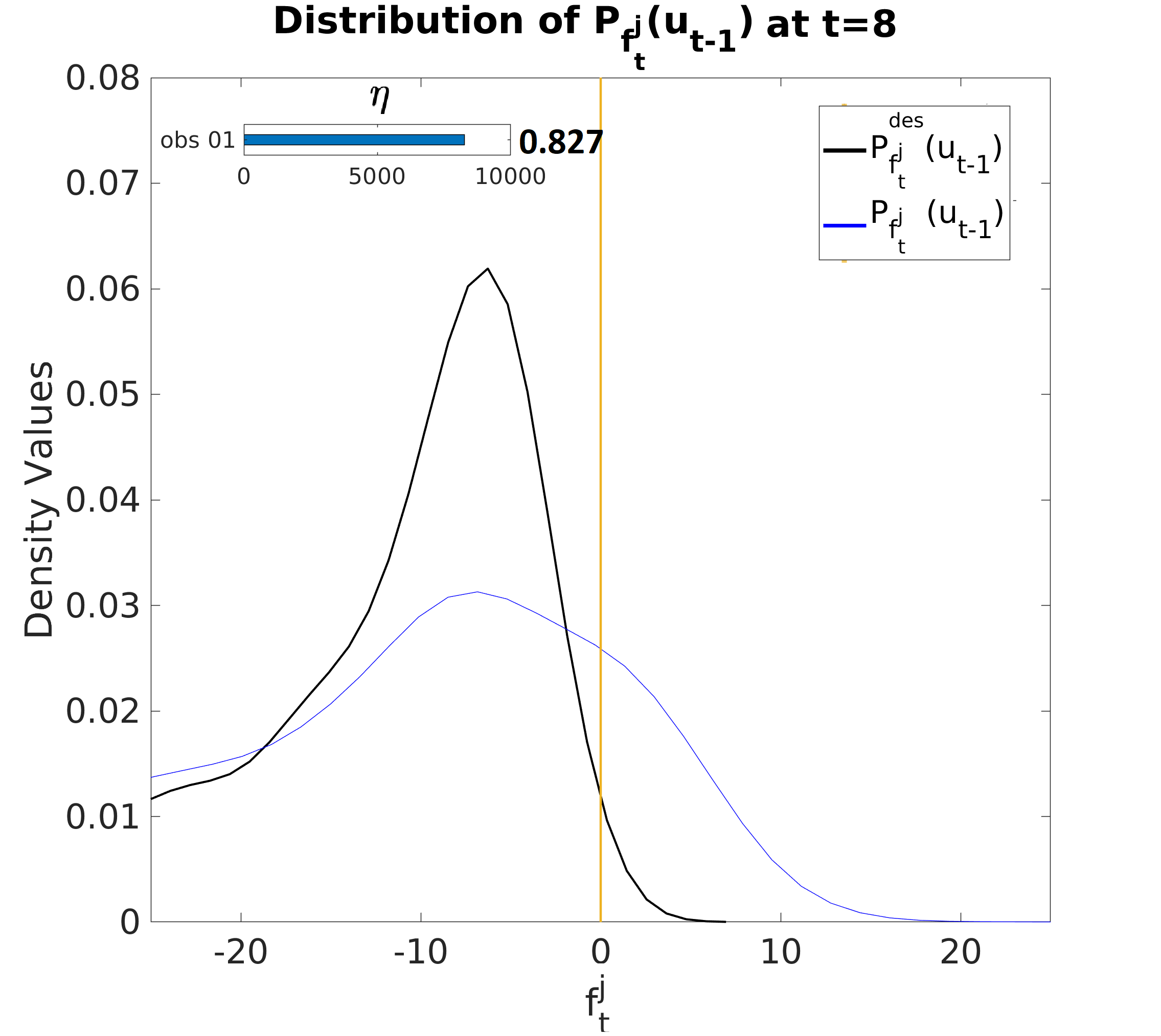}\hfill
        \caption{d = 1}
        \label{Abl2a}
    \end{subfigure}
    \begin{subfigure}{.33\textwidth}
        \includegraphics[width=0.75\linewidth]{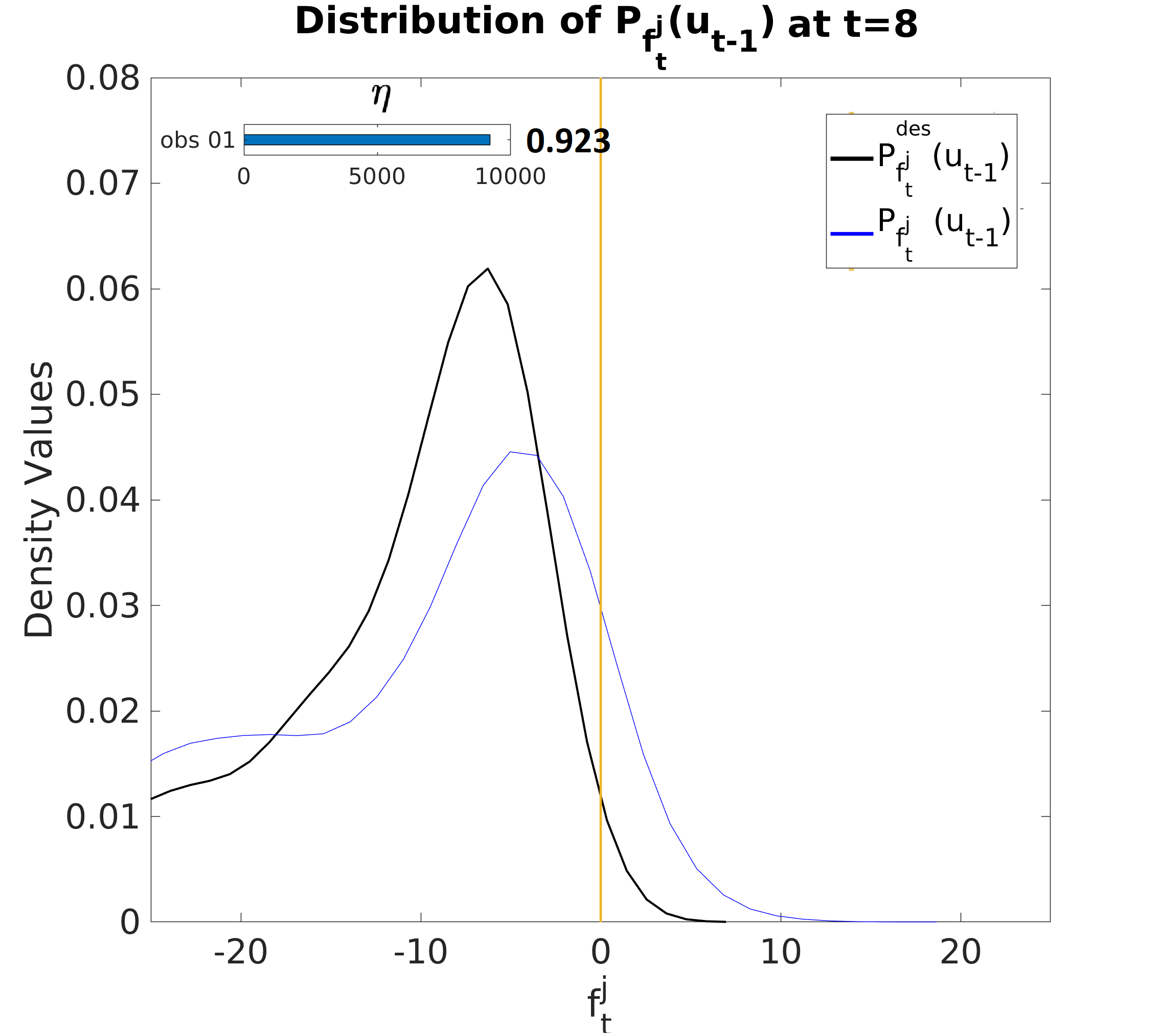}\hfill
        \caption{d = 2}
        \label{Abl2b}
    \end{subfigure}
    \begin{subfigure}{.33\textwidth}
        \includegraphics[width=0.75\linewidth]{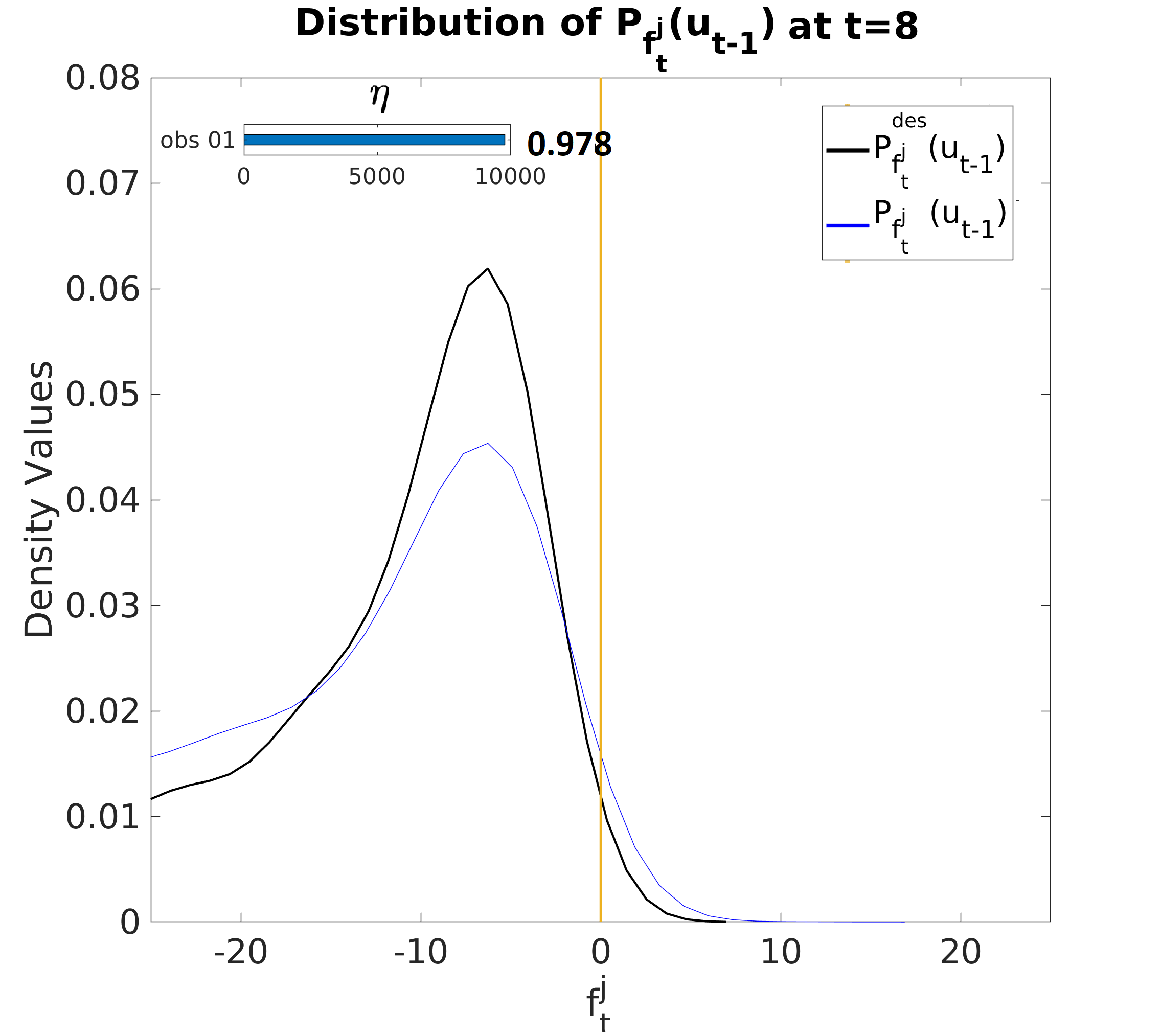}\hfill
        \caption{d = 3}
        \label{Abl2c}
    \end{subfigure}
    \caption{\small \ref{Abl1a}, \ref{Abl1b}, \ref{Abl1c} are the snapshots of the collision avoidance simulation for d = 1,2,3 respectively. An increase in d results in an increase in the  clearance between the robot and the obstacles. The increased clearance in turn results in an improvement to the probability of collision avoidance. \ref{Abl2a}, \ref{Abl2b}, \ref{Abl2c} are the distributions of velocity obstacles $P(f^j_t(\textbf{u}_t))$ obtained after solving the RKHS based approach for different values of d. We see that as d increases more of the tail end starts to match which is also shown by the increasing values of $\eta$.\normalsize}
    \label{Abl}
\end{figure*}

\begin{figure*}
    \begin{subfigure}{.33\textwidth}
        \includegraphics[width=.9\linewidth]{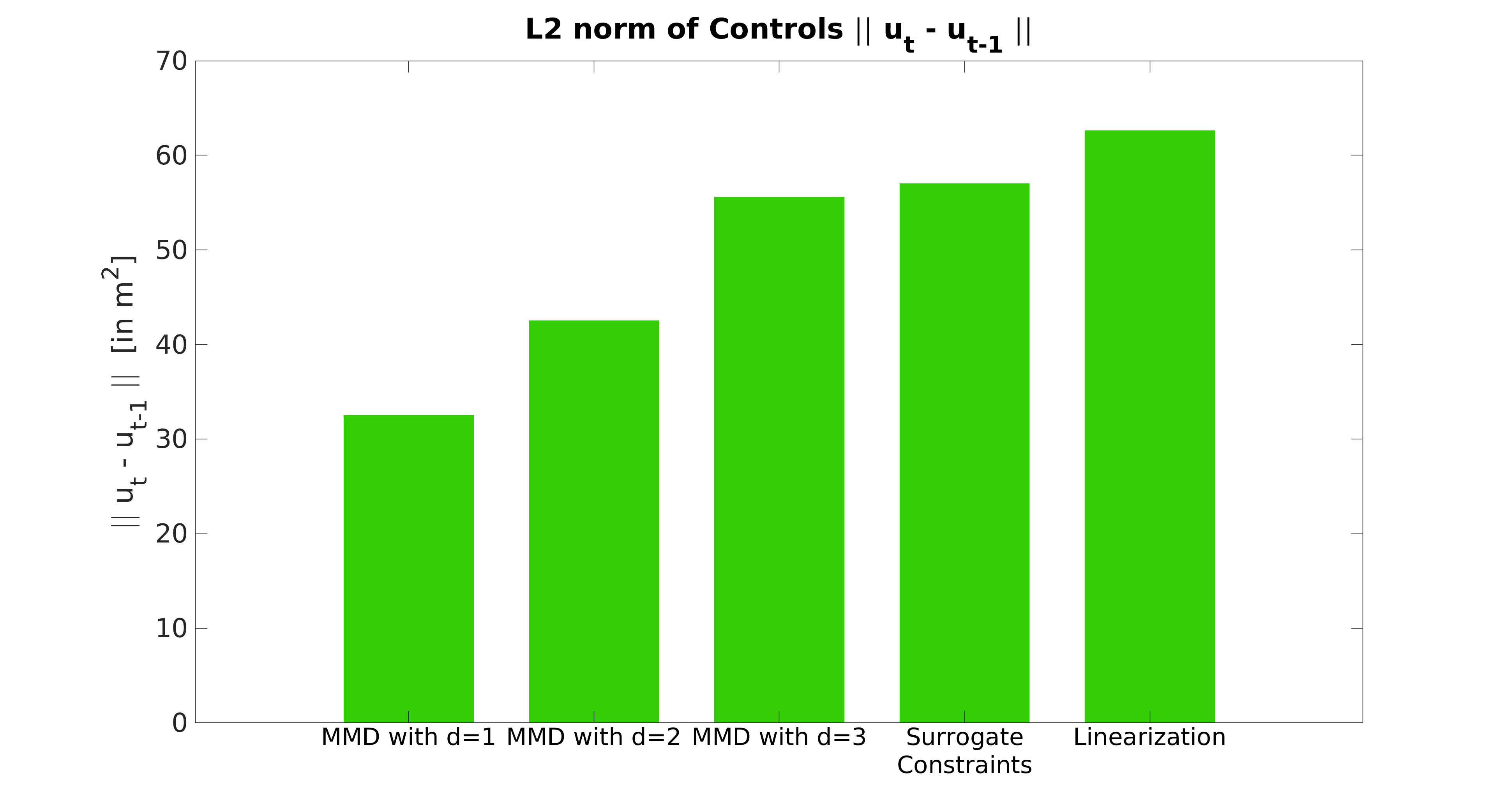}
        \caption{Control costs}
        \label{Bar1}
    \end{subfigure}
    \begin{subfigure}{.33\textwidth}
        \includegraphics[width=.9\linewidth]{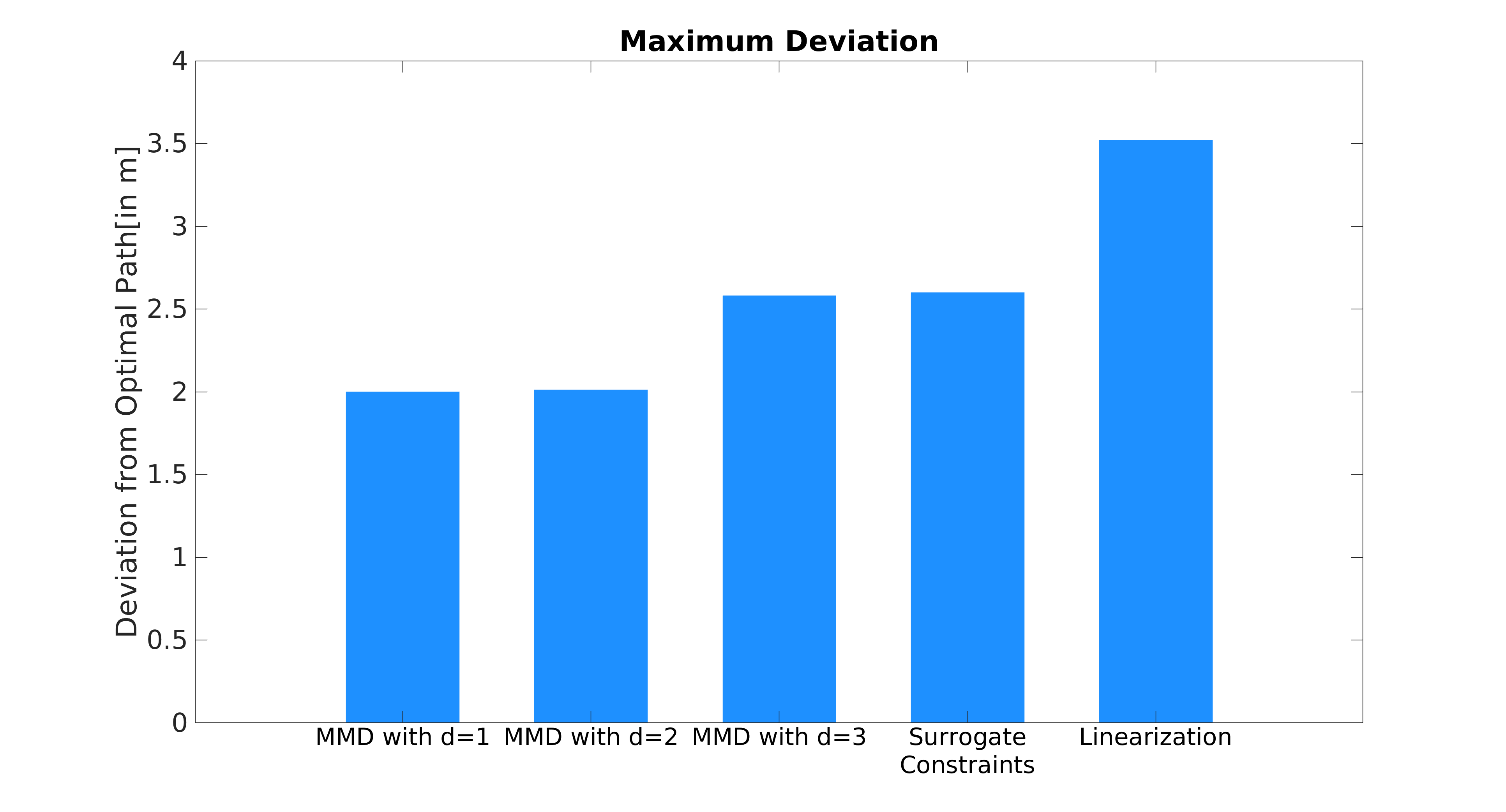}
        \caption{Deviation plots}
        \label{Bar2}
    \end{subfigure}    
        \begin{subfigure}{.33\textwidth}
        \includegraphics[width=.9\linewidth]{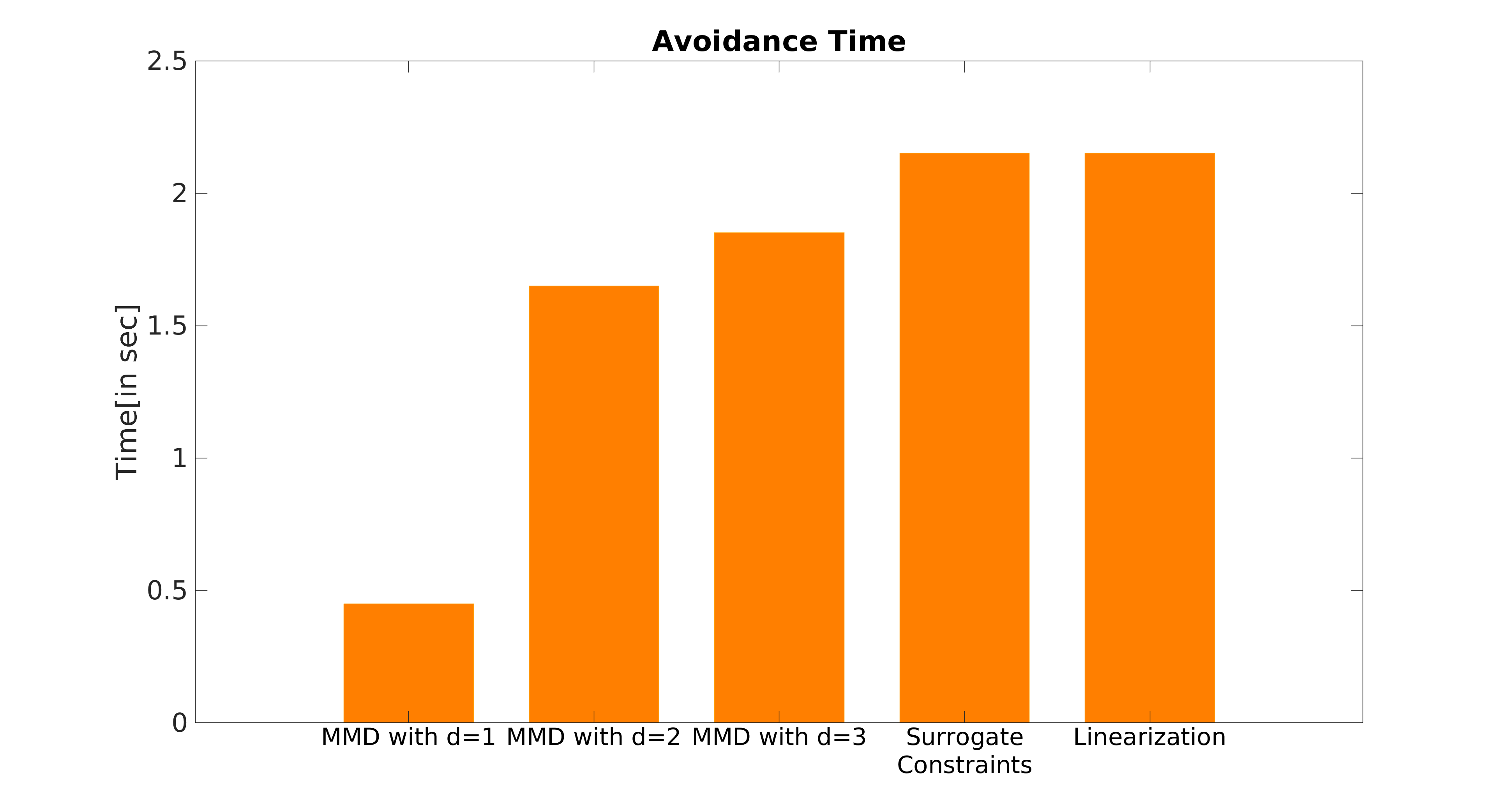}
        \caption{Time duration}
        \label{Bar3}
    \end{subfigure}
    \caption{\small \ref{Bar1} shows the average control cost \ref{Bar2} shows the average deviation from the optimal trajectory \ref{Bar3} shows the time taken to avoid. All costs were obtained with different methods for collision avoidance of a single obstacle observed across n different instances. The proposed RKHS formulation consistently results in a lower cost more optimal solution compared to the listed baseline approaches. Furthermore, the approach based on surrogate constraints is often infeasible for higher $\eta$. \normalsize}
    \label{Bar plots}
\end{figure*}

As the RKHS hyper-parameter $d$ increases higher order moments of the distributions strive to match each other as shown in \ref{Abl2a}, \ref{Abl2b}, \ref{Abl2c}. This results in enhanced collision avoidance at the cost of increasing trend in conservativeness of the maneuver. We show this for a robot avoiding single obstacle in Figures \ref{Abl1a} \ref{Abl1b} \ref{Abl1c}. Figure \ref{Abl1a} depicts the scenario for $d=1$ where the agent gets close enough to the obstacle and not all samples of the obstacles are avoided. As $d$ increases to $d=2$, $d=3$ more samples of the obstacles are avoided by the robot samples as seen in Figures \ref{Abl1b}, \ref{Abl1c}. However the conservativeness of the maneuver depicted in terms of deviation from the optimal path (straight line) also increases with increasing $d$ as shown in the bar chart of Figures \ref{Bar plots}. We further compare with the following 2 formulations. In the first one the collision cone distribution is obtained by further linearization of the Gaussian distributions \cite{alonso-mora-ral19}. In the second one the mean and variance are derived from the Gaussian approximation of the noise. Then we construct the surrogate chance constraint along the lines of \cite{iros15_bharath, prvo}. We consistently perform better than them with respect to both the control cost and the deviation in \ref{Bar plots}.

\begin{figure*}
    \begin{subfigure}{.33\textwidth}
        \includegraphics[width=.9\linewidth]{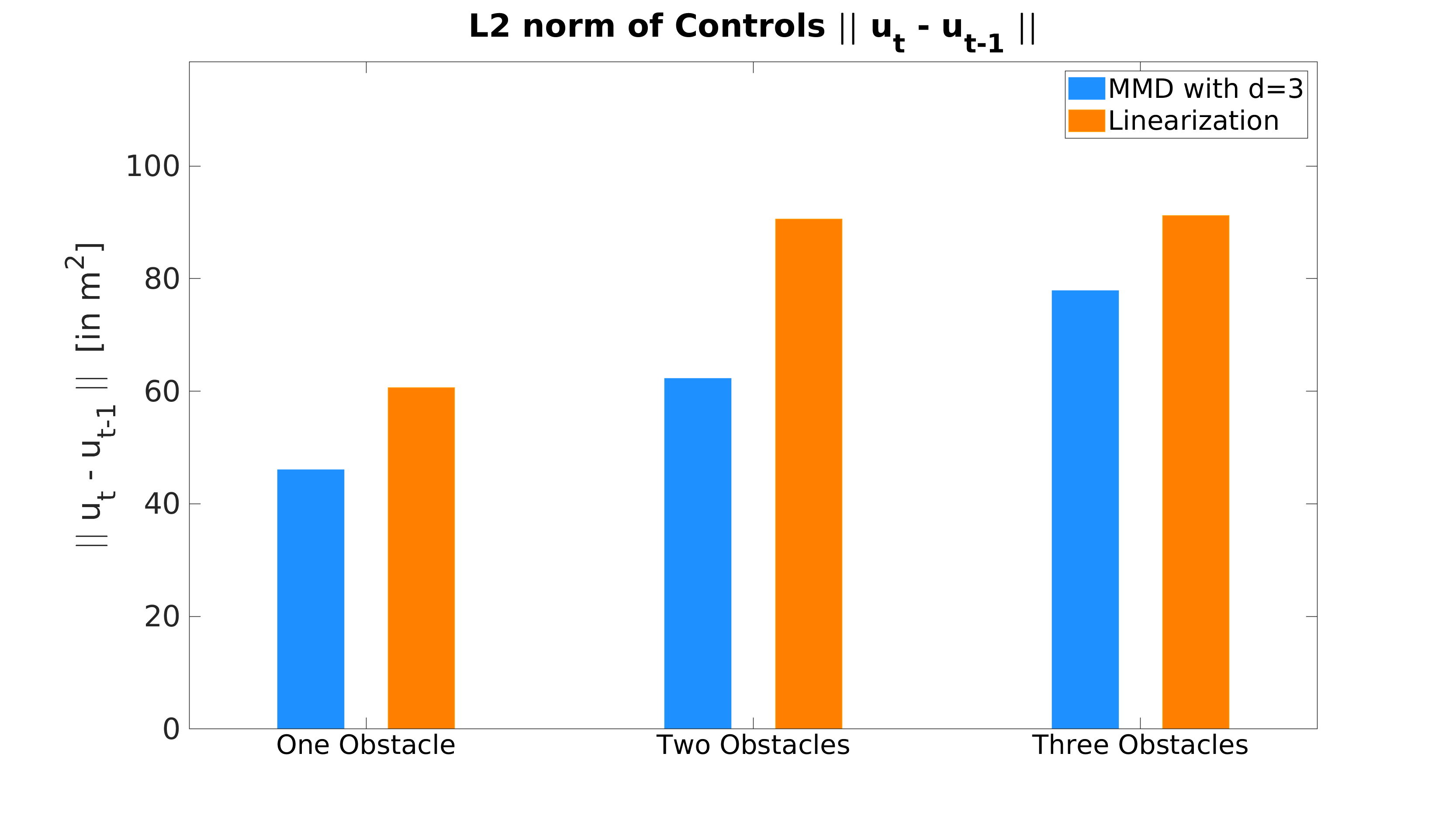}
        \caption{Control costs}
        \label{Multi_Bar1}
    \end{subfigure}
    \begin{subfigure}{.33\textwidth}
        \includegraphics[width=.9\linewidth]{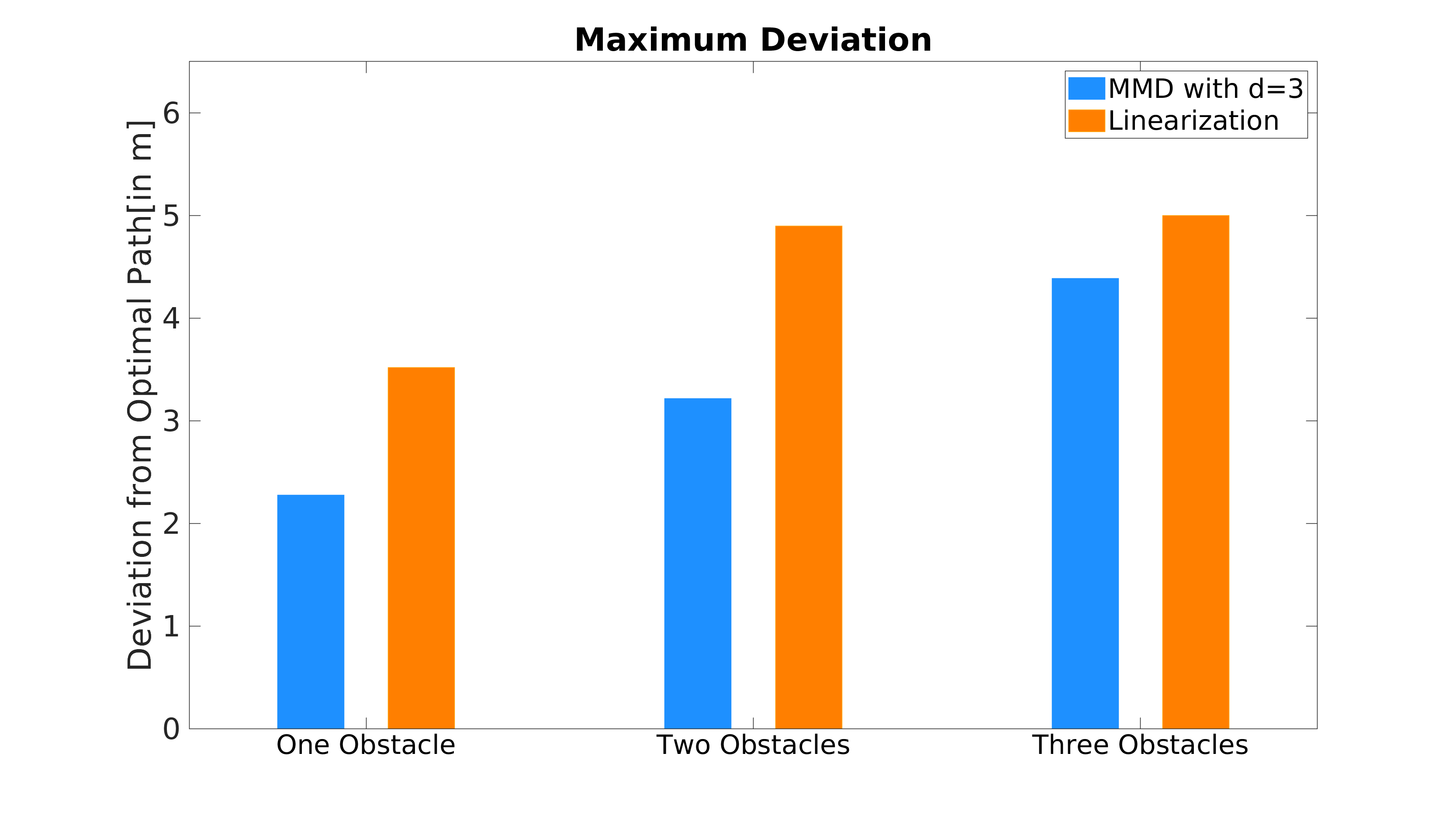}
        \caption{Deviation plots}
        \label{Multi_Bar2}
    \end{subfigure}    
        \begin{subfigure}{.33\textwidth}
        \includegraphics[width=.9\linewidth]{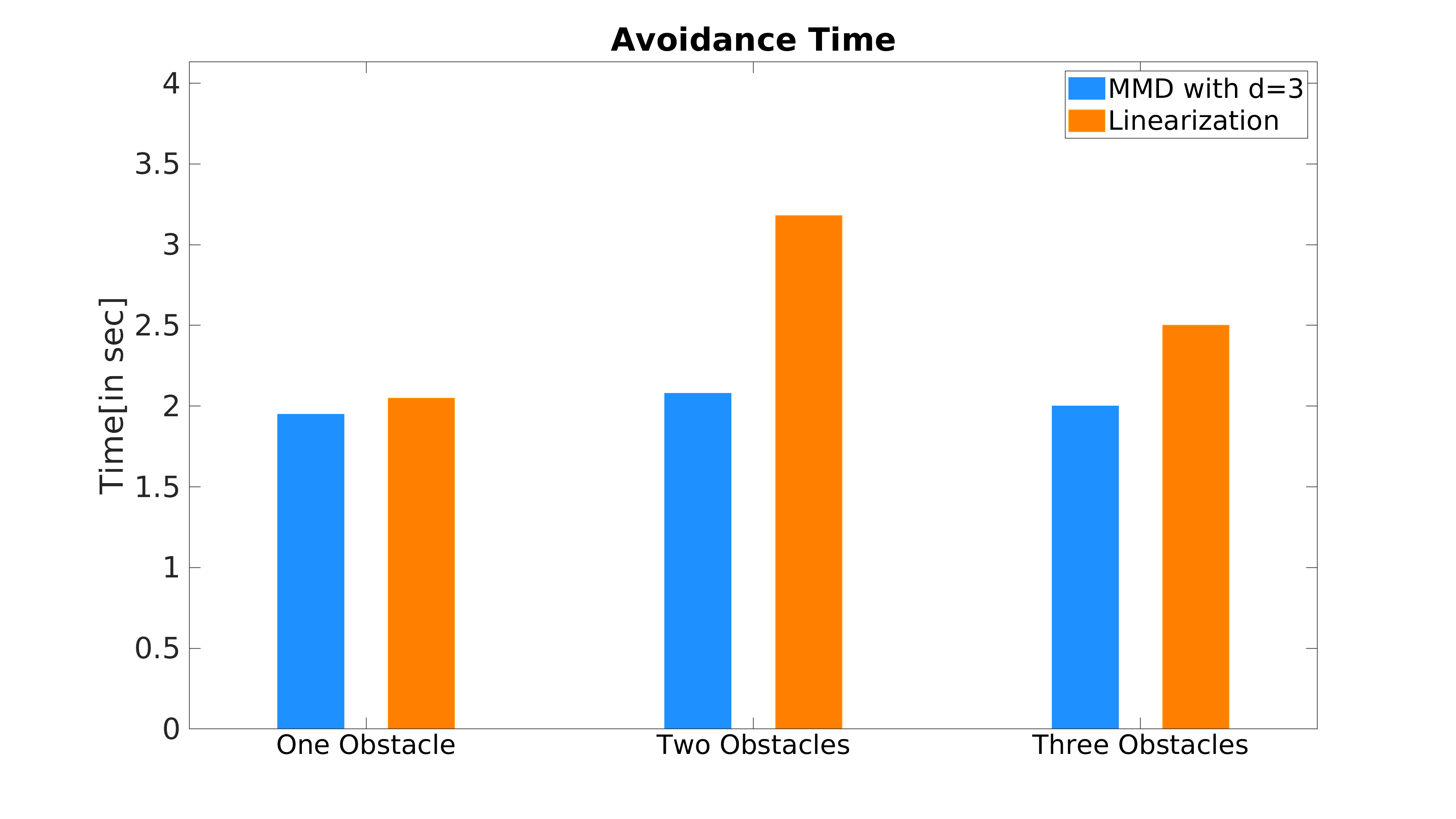}
        \caption{Time duration}
        \label{Multi_Bar3}
    \end{subfigure}
    \caption{\small \ref{Multi_Bar1} shows the average control cost \ref{Multi_Bar2} shows the average deviation from the optimal trajectory \ref{Multi_Bar3} shows the time taken to avoid. All costs were obtained with different methods for collision avoidance of the given number of obstacle observed across n different instances.\normalsize}
    \label{Multi Bar plots}
\end{figure*}

\subsection{Quantitative Comparisons}
\label{sec:multiobs}
    In this section, we compare the proposed RKHS formulation vis a vis with a chance constrained linearized formulation which approximates the non-parametric distribution by Gaussian models as described in Section \ref{sec:Ablation}. The comparisons are shown in the bar plots of Figure \ref{Multi_Bar2} which compares over deviations in trajectory length from the optimal trajectory whereas Figure \ref{Multi_Bar1} compares over the L2 norm of control or actuator costs. The bar coloured in blue in each bar plot  represents the proposed formulation for $d=3$, while the bar coloured in orange is for the method of linearization. Figures \ref{Multi_Bar1}, \ref{Multi_Bar2} and \ref{Multi_Bar3} show the respective comparisons for the cases of one, two and three obstacles between the proposed method and the linearized method.

It is evident from these plots that the RKHS formulation outperforms our baseline on both trajectory deviations and control costs.
Additional results based on the proposed method can be found at \href{https://robotics.iiit.ac.in/publications/2020/non-holonomic-collision-avoidance-hilbert-space.html}{https://robotics.iiit.ac.in/publications/2020/non-holonomic-collision-avoidance-hilbert-space.html}.

%% file: chapters/conclusion.tex
\section{Conclusion}
This paper proposed a novel formulation based on Reproducing Kernel Hilbert Space embedding of non-parametric agent and obstacle distributions and gainfully employed it to solve the problem of the ego agent avoiding several dynamic obstacles under the duress of non-parametric state, velocity, lane boundary, and actuator noise. The formulation was validated for a kinematically constrained agent whose state evolves according to unicycle kinematics. Performance gain across various parameters such as trajectory length and time were shown with respect to two formulations first where the collision cone distribution is obtained by further linearization of the Gaussian distributions with respect to state and velocity variables and the second method where the collision cone distribution is not analytically computed but its mean and variance are derived from the Gaussian approximation which is then  used to construct the surrogate chance constraint. To the best of the authors' knowledge, this is the first such paper to have attacked this problem.

%% file: root.bbl
\begin{thebibliography}{10}
\providecommand{\url}[1]{#1}
\csname url@samestyle\endcsname
\providecommand{\newblock}{\relax}
\providecommand{\bibinfo}[2]{#2}
\providecommand{\BIBentrySTDinterwordspacing}{\spaceskip=0pt\relax}
\providecommand{\BIBentryALTinterwordstretchfactor}{4}
\providecommand{\BIBentryALTinterwordspacing}{\spaceskip=\fontdimen2\font plus
\BIBentryALTinterwordstretchfactor\fontdimen3\font minus
  \fontdimen4\font\relax}
\providecommand{\BIBforeignlanguage}[2]{{%
\expandafter\ifx\csname l@#1\endcsname\relax
\typeout{** WARNING: IEEEtran.bst: No hyphenation pattern has been}%
\typeout{** loaded for the language `#1'. Using the pattern for}%
\typeout{** the default language instead.}%
\else
\language=\csname l@#1\endcsname
\fi
#2}}
\providecommand{\BIBdecl}{\relax}
\BIBdecl

\bibitem{Mora-ICRA14}
J.~Alonso-Mora, P.~Gohl, S.~Watson, R.~Siegwart, and P.~Beardsley, ``Shared
  control of autonomous vehicles based on velocity space optimization,'' in
  \emph{2014 IEEE International Conference on Robotics and Automation
  (ICRA)}.\hskip 1em plus 0.5em minus 0.4em\relax IEEE, 2014, pp. 1639--1645.

\bibitem{Manocha-GVO}
D.~Wilkie, J.~Van Den~Berg, and D.~Manocha, ``Generalized velocity obstacles,''
  in \emph{2009 IEEE/RSJ International Conference on Intelligent Robots and
  Systems}.\hskip 1em plus 0.5em minus 0.4em\relax IEEE, 2009, pp. 5573--5578.

\bibitem{pvo-ral2020}
S.~N.~J. {Poonganam}, B.~{Gopalakrishnan}, V.~S. S. B.~K. {Avula}, A.~K.
  {Singh}, K.~M. {Krishna}, and D.~{Manocha}, ``Reactive navigation under
  non-parametric uncertainty through hilbert space embedding of probabilistic
  velocity obstacles,'' \emph{IEEE Robotics and Automation Letters}, vol.~5,
  no.~2, pp. 2690--2697, April 2020.

\bibitem{gopalakrishnan2018tcst}
B.~Gopalakrishnan, A.~K. Singh, K.~M. Krishna, and D.~Manocha, ``Solving chance
  constrained optimization under non-parametric uncertainty through hilbert
  space embedding,'' 2018.

\bibitem{unicycle}
A.~De~Luca, G.~Oriolo, and C.~Samson, ``Feedback control of a nonholonomic
  car-like robot,'' in \emph{Robot motion, planning and control}, ser. Lecture
  Notes in Control and Information Sciences 229, J.-P. Laumond, Ed.\hskip 1em
  plus 0.5em minus 0.4em\relax Springer, 1998.

\bibitem{vo}
P.~Fiorini and Z.~Shiller, ``Motion planning in dynamic environments using
  velocity obstacles,'' \emph{The International Journal of Robotics Research},
  vol.~17, no.~7, pp. 760--772, 1998.

\bibitem{manocha-rvo}
J.~{van den Berg}, {Ming Lin}, and D.~{Manocha}, ``Reciprocal velocity
  obstacles for real-time multi-agent navigation,'' in \emph{2008 IEEE
  International Conference on Robotics and Automation}, May 2008, pp.
  1928--1935.

\bibitem{prvo}
B.~Gopalakrishnan, A.~K. Singh, M.~Kaushik, K.~M. Krishna, and D.~Manocha,
  ``Prvo: Probabilistic reciprocal velocity obstacle for multi robot navigation
  under uncertainty,'' in \emph{Intelligent Robots and Systems (IROS), 2017
  IEEE/RSJ International Conference on}.\hskip 1em plus 0.5em minus 0.4em\relax
  IEEE, 2017, pp. 1089--1096.

\bibitem{scholkopf}
B.~Sch{\"o}lkopf, K.~Muandet, K.~Fukumizu, S.~Harmeling, and J.~Peters,
  ``Computing functions of random variables via reproducing kernel hilbert
  space representations,'' \emph{Statistics and Computing}, vol.~25, no.~4, pp.
  755--766, 2015.

\bibitem{scholkopf2}
A.~Scibior, C.-J. Simon-Gabriel, I.~O. Tolstikhin, and B.~Sch{\"o}lkopf,
  ``Consistent kernel mean estimation for functions of random variables,'' in
  \emph{Advances in Neural Information Processing Systems}, 2016, pp.
  1732--1740.

\bibitem{iros15_bharath}
B.~Gopalakrishnan, A.~K. Singh, and K.~M. Krishna, ``Closed form
  characterization of collision free velocities and confidence bounds for
  non-holonomic robots in uncertain dynamic environments,'' in
  \emph{Intelligent Robots and Systems (IROS), 2015 IEEE/RSJ International
  Conference on}.\hskip 1em plus 0.5em minus 0.4em\relax IEEE, 2015, pp.
  4961--4968.

\bibitem{bounding_volume1}
D.~Hennes, D.~Claes, W.~Meeussen, and K.~Tuyls, ``Multi-robot collision
  avoidance with localization uncertainty,'' in \emph{Proceedings of the 11th
  International Conference on Autonomous Agents and Multiagent Systems-Volume
  1}.\hskip 1em plus 0.5em minus 0.4em\relax International Foundation for
  Autonomous Agents and Multiagent Systems, 2012, pp. 147--154.

\bibitem{bounding_volume2}
J.~Snape, J.~Van Den~Berg, S.~J. Guy, and D.~Manocha, ``The hybrid reciprocal
  velocity obstacle,'' \emph{IEEE Transactions on Robotics}, vol.~27, no.~4,
  pp. 696--706, 2011.

\bibitem{boyd_chance}
S.~Boyd, ``Stochastic programming,'' \emph{Lecture Notes, Stanford University},
  2008.

\bibitem{chance_blackmore}
L.~Blackmore, M.~Ono, A.~Bektassov, and B.~C. Williams, ``A probabilistic
  particle-control approximation of chance-constrained stochastic predictive
  control,'' \emph{IEEE transactions on Robotics}, vol.~26, no.~3, pp.
  502--517, 2010.

\bibitem{Bratz}
A.~Mesbah, S.~Streif, R.~Findeisen, and R.~Braatz, ``Stochastic nonlinear model
  predictive control with probabilistic constraints,'' 06 2014, pp. 2413--2419.

\bibitem{Kothari}
B.~Luders, M.~Kothari, and J.~How, ``Chance constrained rrt for probabilistic
  robustness to environmental uncertainty,'' in \emph{AIAA guidance,
  navigation, and control conference}, 2010, p. 8160.

\bibitem{moments-determine-tail}
\BIBentryALTinterwordspacing
B.~G. Lindsay and P.~Basak, ``Moments determine the tail of a distribution (but
  not much else),'' \emph{The American Statistician}, vol.~54, no.~4, pp.
  248--251, 2000. [Online]. Available:
  \url{http://www.jstor.org/stable/2685775}
\BIBentrySTDinterwordspacing

\bibitem{kme-nips}
C.-J. Simon-Gabriel, A.~Scibior, I.~O. Tolstikhin, and B.~Sch\"{o}lkopf,
  ``Consistent kernel mean estimation for functions of random variables,'' in
  \emph{Advances in Neural Information Processing Systems 29}, D.~D. Lee,
  M.~Sugiyama, U.~V. Luxburg, I.~Guyon, and R.~Garnett, Eds.\hskip 1em plus
  0.5em minus 0.4em\relax Curran Associates, Inc., 2016, pp. 1732--1740.

\bibitem{alonso-mora-ral19}
H.~{Zhu} and J.~{Alonso-Mora}, ``Chance-constrained collision avoidance for
  mavs in dynamic environments,'' \emph{IEEE Robotics and Automation Letters},
  vol.~4, no.~2, pp. 776--783, April 2019.

\end{thebibliography}
